\documentclass[10pt]{article} 
\usepackage[accepted]{tmlr}

\usepackage{amsmath,amsfonts,bm}

\def\eqref#1{equation~\ref{#1}}

\def\1{\bm{1}}

\DeclareMathAlphabet{\mathsfit}{\encodingdefault}{\sfdefault}{m}{sl}
\SetMathAlphabet{\mathsfit}{bold}{\encodingdefault}{\sfdefault}{bx}{n}

\usepackage{hyperref}
\usepackage{url}
\usepackage{algorithm}
\usepackage{algpseudocode}

\usepackage{microtype}
\usepackage{graphicx}
\usepackage{subfigure}
\usepackage{booktabs} 
\newcommand{\fix}[1]{\textcolor{black}{#1}}
\usepackage{pgfplots}
\usetikzlibrary{pgfplots.groupplots}
\pgfplotsset{compat=1.3}
\usepackage{tikz}
\usetikzlibrary{patterns}
\usepackage{url}            
\usepackage{booktabs}       
\usepackage{amsfonts}       
\usepackage{nicefrac}       
\usepackage{microtype}      
\usepackage{xcolor}         
\usepackage{amsmath}
\usepackage{amssymb}
\usepackage{mathtools}
\usepackage{amsthm}
\usepackage{multirow}
\usepackage{pgfplots}
\usepackage{wrapfig, lipsum, booktabs}

\usepackage{pifont} 
\newcommand{\okmark}{{\textbf{\textcolor[rgb]{0.1, 0.5, 0.1}{$\checkmark$}}}}
\newcommand{\ngmark}{{\textbf{\color{red}{\ding{55}}}}}

\definecolor{battleshipgrey}{rgb}{0.3, 0.3, 0.3}
\definecolor{brilliantrose}{rgb}{1.0, 0.33, 0.64}
\definecolor{americanrose}{rgb}{1.0, 0.01, 0.24}
\definecolor{jweigreen}{rgb}{0,0.45,0.24}
\definecolor{bluegray}{rgb}{0.1, 0.1, 0.4}
\definecolor{ao(english)}{rgb}{0.0, 0.5, 0.0}
\definecolor{blanchedalmond}{rgb}{1.0, 0.92, 0.8}
\definecolor{atomictangerine}{rgb}{1.0, 0.6, 0.4}
\definecolor{chocolate(web)}{rgb}{0.82, 0.41, 0.12}
\definecolor{bananayellow}{rgb}{1.0, 0.88, 0.21}
\definecolor{goldenbrown}{rgb}{0.6, 0.4, 0.08}
\definecolor{aliceblue}{rgb}{0.94, 0.97, 1.0}
\definecolor{beige}{rgb}{0.96, 0.96, 0.86}
\definecolor{babyblue}{rgb}{0.54, 0.81, 0.94}
\definecolor{camel}{rgb}{0.76, 0.6, 0.42}
\definecolor{cinnamon}{rgb}{0.82, 0.41, 0.12}

\usepackage{hyperref}
\usepackage{url}

\title{Multimodal Chain-of-Thought Reasoning in \\ Language Models}

\author{\name Zhuosheng Zhang\thanks{Work done at Amazon Web Services. Correspondence to: Zhuosheng Zhang and Aston Zhang.} \email zhangzs@sjtu.edu.cn \\
      \addr School of Electronic Information and Electrical Engineering, \\Shanghai Jiao Tong University
      \AND
      \name Aston Zhang$^{*}$ \email az@astonzhang.com \\
      \addr GenAI, Meta 
      \AND
      \name Mu Li \email muli@cs.cmu.edu \\
      \addr Amazon Web Services
      \AND
      \name Hai Zhao \email zhaohai@cs.sjtu.edu.cn \\
      \addr Department of Computer Science and Engineering, \\Shanghai Jiao Tong University 
      \AND
      \name George Karypis \email gkarypis@amazon.com \\
      \addr Amazon Web Services
      \AND
      \name Alex Smola \email alex@smola.org \\
      \addr Amazon Web Services
      }

\def\month{05}  
\def\year{2024} 
\def\openreview{\url{https://openreview.net/forum?id=y1pPWFVfvR}} 

\begin{document}

\maketitle

\begin{abstract}
Large language models (LLMs) have shown impressive performance on complex reasoning by leveraging chain-of-thought (CoT) prompting to generate intermediate reasoning chains as the rationale to infer the answer. However, existing CoT studies have primarily focused on the language modality. We propose Multimodal-CoT that incorporates language (text) and vision (images) modalities into a two-stage framework that separates rationale generation and answer inference. 
In this way, answer inference can leverage better generated rationales that are based on multimodal information. Experimental results on ScienceQA and A-OKVQA benchmark datasets show the effectiveness of our proposed approach. With Multimodal-CoT, our model under 1 billion parameters achieves state-of-the-art performance on the ScienceQA benchmark. Our analysis indicates that Multimodal-CoT offers the advantages of mitigating hallucination and enhancing convergence speed. Code is publicly available at \texttt{https://github.com/amazon-science/mm-cot}.
\end{abstract}

\section{Introduction}
\label{intro}
Imagine reading a textbook with no figures or tables. Our ability to knowledge acquisition is greatly strengthened by jointly modeling diverse data modalities, such as vision, language, and audio. Recently, large language models (LLMs) \citep{brown2020language, lamda, gopher, palm} have shown impressive performance in complex reasoning by generating intermediate reasoning steps before inferring the answer. The intriguing technique is called chain-of-thought (CoT) reasoning \citep{cot_wei,kojima2022large,zhang2022automatic}. 

\begin{wrapfigure}{r}{0.5\textwidth}
\centering
\vspace{-0.5mm}
\includegraphics[width=0.5\textwidth]{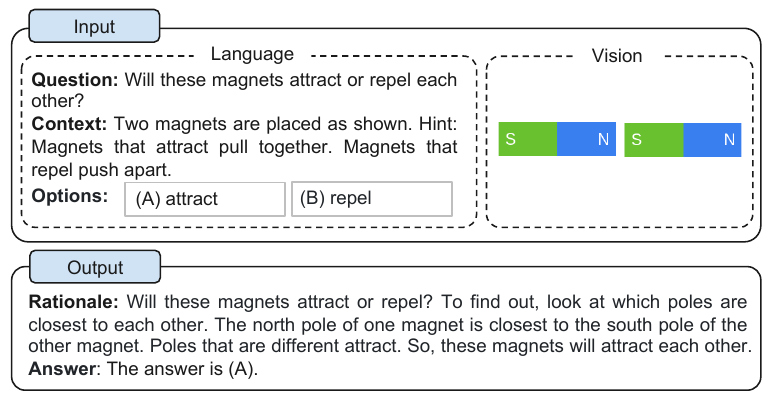}
  \vspace{-6.8mm}
  \caption{Example of the multimodal CoT task.}
  \vspace{-6mm}
\label{fig_examples}
\end{wrapfigure}



However, existing studies related to CoT reasoning are largely isolated in the language modality \citep{wang2022rationale,zhou2022least,lu2022dynamic,fu2022complexity}, with little consideration of multimodal scenarios. To elicit CoT reasoning in multimodality, we advocate a Multimodal-CoT paradigm. Given the inputs in different modalities, Multimodal-CoT decomposes multi-step problems into intermediate reasoning steps (rationale) and then infers the answer. Since vision and language are the most popular modalities, we focus on those two modalities in this work. An example is shown in Figure \ref{fig_examples}. 


In general, Multimodal-CoT reasoning can be elicited through two primary paradigms: (i) prompting LLMs and (ii) fine-tuning smaller models.\footnote{We refer to small models as models with less than 1 billion parameters (hereinafter dubbed as 1B-models).} We will delve into these paradigms and delineate their associated challenges as follows.


The most immediate way to perform Multimodal-CoT is to transform the input of different modalities into a unified modality and prompt LLMs to perform CoT \citep{zhang2023llama,lu2023chameleon,liu2023visual,alayrac2022flamingo,hao2022language,yasunaga2022retrieval}. {For example}, it is possible to generate a caption for an image by a captioning model and then concatenate the caption with the original language input to be fed into LLMs \citep{lu2022learn}. 
{The development of large multimodal models such as GPT-4V \citep{gpt4v} and Gemini \citep{reid2024gemini} has notably enhanced the quality of generated captions, resulting in finer-grained and more detailed descriptions.
However, the captioning process still incurs significant information loss when transforming vision signals into textual descriptions. Consequently, using image captions rather than vision features may suffer from a lack of mutual synergy in the representation space of different modalities.} In addition, LLMs either have paywalls or resource-consuming to deploy locally.

{To facilitate the interaction between modalities, another potential solution is to fine-tune smaller language models (LMs) by fusing multimodal features \citep{zhang2023universal,zhao2023mmicl}. As this approach allows the flexibility of adjusting model architectures to incorporate multimodal features, we study fine-tuning models in this work instead of prompting LLMs. The key challenge is that language models under 100 billion parameters tend to generate hallucinated rationales that mislead the answer inference \citep{ho2022large,magister2022teaching,ji2022survey,zhang2023siren}.}


{To mitigate the challenge of hallucination, we propose Multimodal-CoT that incorporates language (text) and vision (images) modalities into a two-stage framework that separates rationale generation and answer inference.\footnote{{This work focuses on the language and vision modalities.}} 
In this way, answer inference can leverage better generated rationales that are based on multimodal information.} 
Our experiments were conducted on the  ScienceQA \citep{lu2022learn} and A-OKVQA \citep{schwenk2022okvqa} datasets, which are the latest multimodal reasoning benchmarks with annotated reasoning chains. 

Our method achieves state-of-the-art performance on the ScienceQA benchmark upon the release. We find that Multimodal-CoT is beneficial in mitigating hallucination and boosting convergence. Our contributions are summarized as follows:

(i) {To the best of our knowledge, this work is the first to study CoT reasoning in different modalities in scientific peer-reviewed literature.}

(ii) We propose a two-stage framework by fine-tuning language models to fuse vision and language representations to perform Multimodal-CoT. The model is able to generate informative rationales to facilitate inferring final answers.

(iii) We elicit the analysis of why the naive way of employing CoT fails in the context and how incorporating vision features alleviates the problem. The approach has been shown to be generally effective across tasks and backbone models.






\section{Background}  

This section reviews studies eliciting CoT reasoning by prompting and fine-tuning language models.

\subsection{CoT Reasoning with LLMs}
Recently, CoT has been widely used to elicit the multi-step reasoning abilities of LLMs \citep{cot_wei}. Concretely, CoT techniques encourage the LLM to generate intermediate reasoning chains for solving a problem. Studies have shown that LLMs can perform CoT reasoning with two major paradigms of techniques: Zero-Shot-CoT \citep{kojima2022large} and Few-Shot-CoT \citep{cot_wei,zhang2022automatic}. For Zero-Shot-CoT, \citet{kojima2022large} showed that LLMs are decent zero-shot reasoners by adding a prompt like ``Let’s think step by step'' after the test question to invoke CoT reasoning. For Few-Shot-CoT, a few step-by-step reasoning demonstrations are used as conditions for inference. Each demonstration has a question and a reasoning chain that leads to the final answer. The demonstrations are commonly obtained by hand-crafting or automatic generation. These two techniques, hand-crafting and automatic generation are thus referred to as Manual-CoT \citep{cot_wei} and Auto-CoT \citep{zhang2022automatic}. 

With effective demonstrations, Few-Shot-CoT often achieves stronger performance than Zero-Shot-CoT and has attracted more research interest. Therefore, most recent studies focused on how to improve Few-Shot-CoT. Those studies are categorized into two major research lines: (i) optimizing the demonstrations; (ii) optimizing the reasoning chains. Table \ref{tab:cot_methods} compares typical CoT techniques.

\begin{table*}[t]
\centering
\renewcommand\tabcolsep{2pt} 
\small
\caption{Representative CoT techniques (FT: fine-tuning; KD: knowledge distillation). Segment 1: in-context learning techniques; Segment 2: fine-tuning techniques. To the best of our knowledge, our work is the first to study CoT reasoning in different modalities in scientific peer-reviewed literature. Besides, we focus on 1B-models, without relying on the outputs of LLMs. \label{tab:cot_methods}
}
\scalebox{0.93}{
\begin{tabular}{lccccc} 
\toprule
\textbf{Models} & \textbf{Mutimodal}   & \textbf{Model / Engine}  & \textbf{Training}  & \textbf{CoT Role} & \textbf{CoT Source} \\ 
\midrule
Zero-Shot-CoT~\citep{kojima2022large} & \ngmark  & GPT-3.5 (175B)  & ICL  & Reasoning & Template  \\
Few-Shot-CoT~\citep{cot_wei} & \ngmark  & PaLM (540B) & ICL & Reasoning & Hand-crafted \\
Self-Consistency-CoT~\citep{cot_wei_sc} & \ngmark & Codex  (175B) & ICL  & Reasoning& Hand-crafted   \\
Least-to-Most Prompting~\citep{zhou2022least}& \ngmark & Codex (175B)  & ICL & Reasoning & Hand-crafted   \\
Retrieval-CoT~\citep{zhang2022automatic} & \ngmark& GPT-3.5 (175B) & ICL  & Reasoning & Auto-generated  \\
PromptPG-CoT~\citep{lu2022dynamic} & \ngmark &  GPT-3.5 (175B) & ICL & Reasoning  & Hand-crafted   \\
Auto-CoT~\citep{zhang2022automatic} & \ngmark & Codex (175B)  & ICL & Reasoning  & Auto-generated \\
Complexity-CoT~\citep{fu2022complexity}& \ngmark &  GPT-3.5 (175B) & ICL  & Reasoning & Hand-crafted  \\
Few-Shot-PoT~\citep{chen2022program} & \ngmark &  GPT-3.5 (175B) & ICL & Reasoning  & Hand-crafted  \\ 
\midrule
UnifiedQA~\citep{lu2022learn}& \ngmark  & T5 (770M)& FT & Explanation  & Crawled\\
Fine-Tuned T5 XXL~\citep{magister2022teaching}  & \ngmark   & T5 (11B)& KD & Reasoning  & LLM-generated \\
Fine-Tune-CoT \citep{ho2022large} & \ngmark  & GPT-3 (6.7B) & KD & Reasoning  & LLM-generated \\
Multimodal-CoT (our work) & \okmark   & T5 (770M)& FT & Reasoning  & Crawled   \\ 
\bottomrule
\end{tabular}
}
\vspace{-3mm}
\end{table*}

\paragraph{Optimizing Demonstrations} The performance of Few-Shot-CoT relies on the quality of demonstrations. As reported in \citet{cot_wei}, using demonstrations written by different annotators results in dramatic accuracy disparity in reasoning tasks. Beyond hand-crafting the demonstrations, recent studies have investigated ways to optimize the demonstration selection process. Notably, \citet{rubin2021learning} retrieved the semantically similar demonstrations with the test instance. However, this approach shows a degraded performance when there are mistakes in the reasoning chains \citep{zhang2022automatic}. To address the limitation, \citet{zhang2022automatic} found that the key is the diversity of demonstration questions and proposed Auto-CoT: (i) partition questions of a given dataset into a few clusters; (ii) sample a representative question from each cluster and generate its reasoning chain using Zero-Shot-CoT with simple heuristics. In addition, reinforcement learning (RL) and complexity-based selection strategies were proposed to obtain effective demonstrations. \citet{fu2022complexity} chose examples with complex reasoning chains (i.e., with more reasoning steps) as the demonstrations. \citet{lu2022dynamic} trained an agent to find optimal in-context examples from a candidate pool and maximize the prediction rewards on given training examples when interacting with GPT-3.5.

\paragraph{Optimizing Reasoning Chains} A notable way to optimize reasoning chains is problem decomposition.  \citet{zhou2022least} proposed least-to-most prompting to decompose complex problems into sub-problems and then solve these sub-problems sequentially. As a result, solving a given sub-problem is facilitated by the answers to previously solved sub-problems. Similarly, \citet{khot2022decomposed} used diverse decomposition structures and designed different prompts to answer each sub-question. In addition to prompting the reasoning chains as natural language texts, \citet{chen2022program} proposed program-of-thoughts (PoT), which modeled the reasoning process as a program and prompted LLMs to derive the answer by executing the generated programs. Another trend is to vote over multiple reasoning paths for a test question. \citet{cot_wei_sc} introduced a self-consistency decoding strategy to sample multiple outputs of LLMs and then took a majority over the final answers. \citet{wang2022rationale} and \citet{li2022advance} introduced randomness in the input space to produce more diverse outputs for voting.

\subsection{Eliciting CoT Reasoning by Fine-Tuning Models}
A recent interest is eliciting CoT reasoning by fine-tuning language models. \citet{lu2022learn} fine-tuned the encoder-decoder T5 model on a large-scale dataset with CoT annotations. However, a dramatic performance decline is observed when using CoT to infer the answer, i.e., generating the reasoning chain before the answer (reasoning). Instead, CoT is only used as an explanation after the answer. \citet{magister2022teaching} and \citet{ho2022large} employed knowledge distillation by fine-tuning a student model on the chain-of-thought outputs generated by a larger teacher model. \citet{wang2022iteratively} proposed an iterative context-aware prompting approach to dynamically synthesize prompts conditioned on the current step's contexts.

{There is a key challenge in training 1B-models to be CoT reasoners.} As observed by \citet{cot_wei}, models under 100 billion parameters tend to produce illogical CoT that leads to wrong answers. In other words, it might be harder for 1B-models to generate effective CoT than directly generating the answer. It becomes even more challenging in a multimodal setting where answering the question also requires understanding the multimodal inputs. In the following part, we will explore the challenge of Multimodal-CoT and investigate how to perform effective multi-step reasoning.

\section{Challenge of Multimodal-CoT}\label{sec:prelim}
Existing studies have suggested that the CoT reasoning ability may emerge in language models at a certain scale, e.g., over 100 billion parameters \citep{wei2022emergent}. However, it remains an unresolved challenge to elicit such reasoning abilities in 1B-models, let alone in the multimodal scenario. {This work focuses on 1B-models as they can be fine-tuned and deployed with consumer-grade GPUs (e.g., 32G memory).}
In this section, we will investigate why 1B-models fail at CoT reasoning and study how to design an effective approach to overcome the challenge.



\subsection{Towards the Role of CoT}

To begin with, we fine-tune a text-only baseline for CoT reasoning on the ScienceQA benchmark \citep{lu2022learn}. We adopt FLAN-Alpaca$_\texttt{Base}$ as the backbone language model.\footnote{\url{https://github.com/declare-lab/flan-alpaca}. It is a 200M T5 model \citep{raffel2020exploring} fine-tuned on  Stanford Alpaca data \citep{taori2023alpaca}. Implementation details are presented in Section \ref{app-sec:baseline}.} Our task is modeled as a text generation problem, where the model takes the textual information as the input and generates the output sequence
that consists of the rationale and the answer.

\begin{wraptable}{r}{0.48\textwidth}
\centering
\vspace{-7mm}
\caption{Effects of CoT in {the one-stage setting}.\label{tab:pre_position}}
         \setlength{\tabcolsep}{12pt}
         \small
         
        \begin{tabular}{llc}\toprule
         {Method} & {Format} & {Accuracy} \\\midrule
        No-CoT & QCM$\rightarrow$A   & 81.63 \\
        \midrule
        Reasoning &  QCM$\rightarrow$RA   & 69.32\\
        Explanation &  QCM$\rightarrow$AR   & 69.68\\
        \bottomrule
        \end{tabular}
        \vspace{-3mm}
\end{wraptable}

As an example shown in Figure \ref{fig_examples}, the model takes the
concatenation of tokens of the question text (Q), the context text (C), and multiple options (M) as the input. To study the effect of CoT, we compare the performance with three variants: (i) \texttt{No-CoT} which predicts the answer directly (QCM$\rightarrow$A); (ii) \texttt{Reasoning} where answer inference is conditioned to the rationale (QCM$\rightarrow$RA); (iii) \texttt{Explanation} where the rationale is used for explaining the answer inference (QCM$\rightarrow$AR).


\begin{figure*}[t]
  \begin{center}
   \includegraphics[width=1\textwidth]{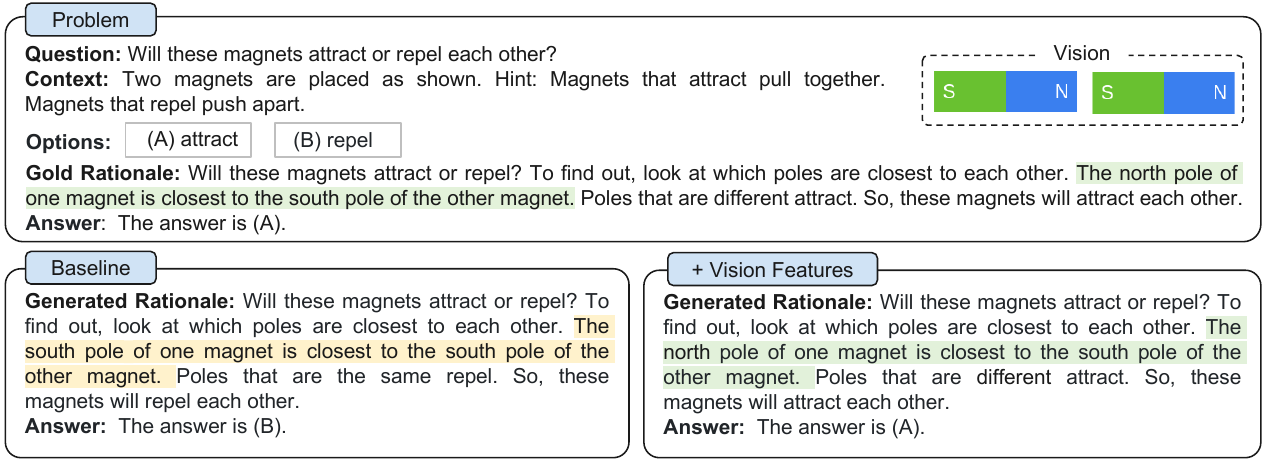}
  \end{center}
  \caption{Example of the two-stage framework without vision features (baseline) and with vision features (ours) for generating rationales and predicting answers. The upper part presents the problem details with a gold rationale, and the lower part shows the outputs of the baseline and our method incorporated with vision features. We observe that the baseline fails to predict the right answer due to the misleading by hallucinated rationales. More examples are shown in Appendix \ref{appendix:misleading}.}
    \vspace{-3mm}
  \label{fig_pre_case1}
\end{figure*}

Surprisingly, as shown in Table \ref{tab:pre_position}, we observe a $\downarrow$12.31\% accuracy decrease (81.63\%$\rightarrow$69.32\%) if the model predicts rationales before answers (QCM$\rightarrow$RA). The results imply that the rationales might not necessarily contribute to predicting the right answer. According to \citet{lu2022learn}, the plausible reason might be that the model exceeds the maximum token limits before obtaining the required answer or stops generating the prediction early. However, we find that the maximum length of the generated outputs (RA) is always less than 400 tokens, which is below the length limit of language models (i.e., 512 in T5 models).
Therefore, it deserves a more in-depth investigation into why the rationales harm answer inference.

\begin{wraptable}{r}{0.48\textwidth}
    \centering\small
    \vspace{-6.8mm}
        \caption{{Two-stage} setting of (i) rationale generation (RougeL) and (ii) answer inference (Accuracy). \label{tab:pre_decoupled}}
    \setlength{\tabcolsep}{0.9pt}
\begin{tabular}{lcc}\toprule
 {Method} & {(i) QCM$ \rightarrow$ R}  & {(ii) QCMR$ \rightarrow$ A}  \\\midrule
 Two-Stage Framework & 90.73 & 78.57 \\
 \midrule
 \quad w/ Captions & 90.88 & 79.37 \\
 \quad w/ Vision Features &  93.46 &	85.31 \\
\bottomrule
\end{tabular}
\vspace{-3mm}
\end{wraptable}

\subsection{Misleading by Hallucinated Rationales}\label{sec:misleading}
To dive into how the rationales affect the answer prediction, we separate the CoT problem into two stages, \textit{rationale generation} and \textit{answer inference}.\footnote{The details will be presented in Section \ref{sec:mm_cot}.} We report the RougeL score and accuracy for the rationale generation and answer inference, respectively. Table \ref{tab:pre_decoupled} shows the results based on the two-stage framework. Although the two-stage baseline model achieves a 90.73 RougeL score of the rationale generation, the answer inference accuracy is only 78.57\%. Compared with the QCM$\rightarrow$A variant (81.63\%) in Table \ref{tab:pre_position}, the result shows that the generated rationale in the two-stage framework does not improve answer accuracy.

Then, we randomly sample 50 error cases and find that the model tends to generate hallucinated rationales that mislead the answer inference. As an example shown in Figure \ref{fig_pre_case1}, the model (left part) hallucinates that, ``\textit{The south pole of one magnet is closest to the south pole of the other magnet}'', due to the lack of reference to the vision content. 
We find that such mistakes occur at a ratio of 56\% among the error cases (Figure \ref{fig_bar}(a)).


\subsection{Multimodality Contributes to Effective Rationales}\label{sec:multimodal}

We speculate that such a phenomenon of hallucination is due to a lack of necessary vision contexts for performing effective Multimodal-CoT. To inject vision information, a simple way is to transform the image into a caption \citep{lu2022learn} and then append the caption in the input of both stages. 
\begin{wrapfigure}{r}{0.48\textwidth}
  \begin{center}
    \vspace{-3mm}
   \includegraphics[width=0.48\textwidth]{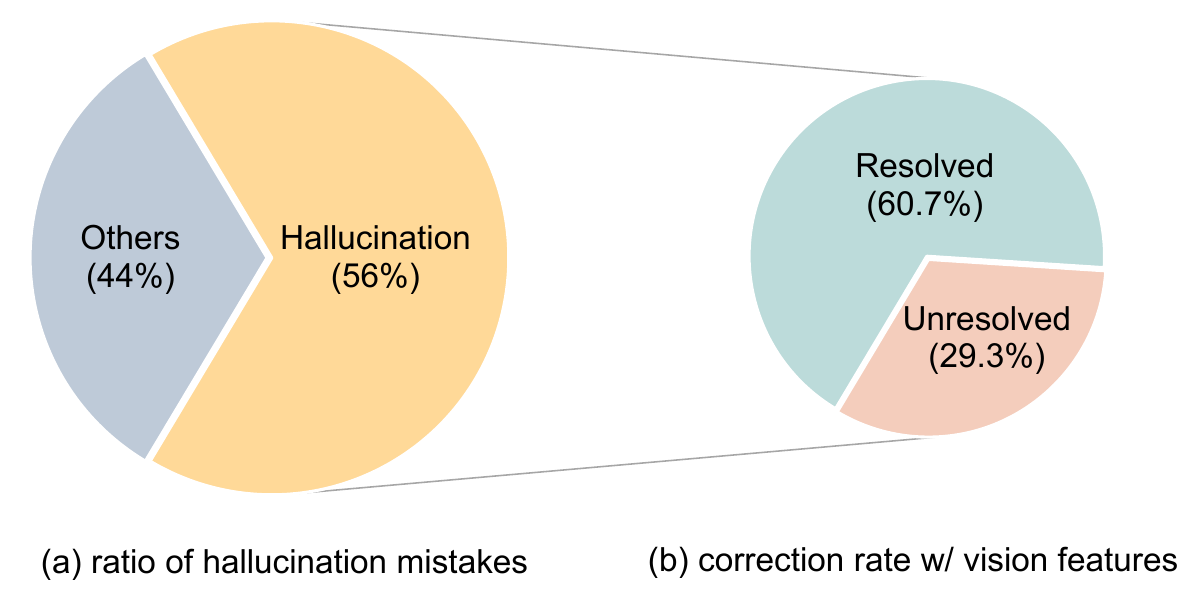}
  \end{center}
  \vspace{-5mm}
  \caption{The ratio of (a) hallucination mistakes and (b) correction rate w/ vision features.}
    \vspace{-3mm}
  \label{fig_bar}
\end{wrapfigure}
However, as shown in Table \ref{tab:pre_decoupled}, using captions only yields marginal performance gains ($\uparrow$0.80\%). Then, we explore an advanced technique by incorporating vision features into the language model. 
Concretely, we feed the image to the ViT model \citep{dosovitskiy2021image} to extract vision features. Then we fuse the vision features with the encoded language representations before feeding the decoder (more details will be presented in Section \ref{sec:mm_cot}). Interestingly, with vision features, the RougeL score of the rationale generation has boosted to 93.46\% (QCM$\rightarrow$R), which correspondingly contributes to better answer accuracy of 85.31\% (QCMR$\rightarrow$A).

With those effective rationales, the phenomenon of hallucination is mitigated --- 60.7\% hallucination mistakes in Section \ref{sec:misleading} have been corrected (Figure \ref{fig_bar}(b)), as an example shown in Figure \ref{fig_pre_case1} (right part).\footnote{The left mistakes are mainly about map understanding, requiring extra commonsense signals (Section \ref{sec:case_studies}).} The analysis so far compellingly shows that vision features are indeed beneficial for generating effective rationales and contributing to accurate answer inference. {As the two-stage method achieves better performance than {one-stage} methods, we choose the two-stage method in our Multimodal-CoT framework.}

\section{Multimodal-CoT}\label{sec:mm_cot}
In light of the discussions in Section \ref{sec:prelim}, {we propose Multimodal-CoT.} {The key motivation is the anticipation that the answer inference can leverage better generated rationales that are based on multimodal information.} In this section, we will overview the procedure of the framework and elaborate on the technical design of the model architecture.

\begin{figure*}[htb]
  \begin{center}
\includegraphics[width=1\textwidth]{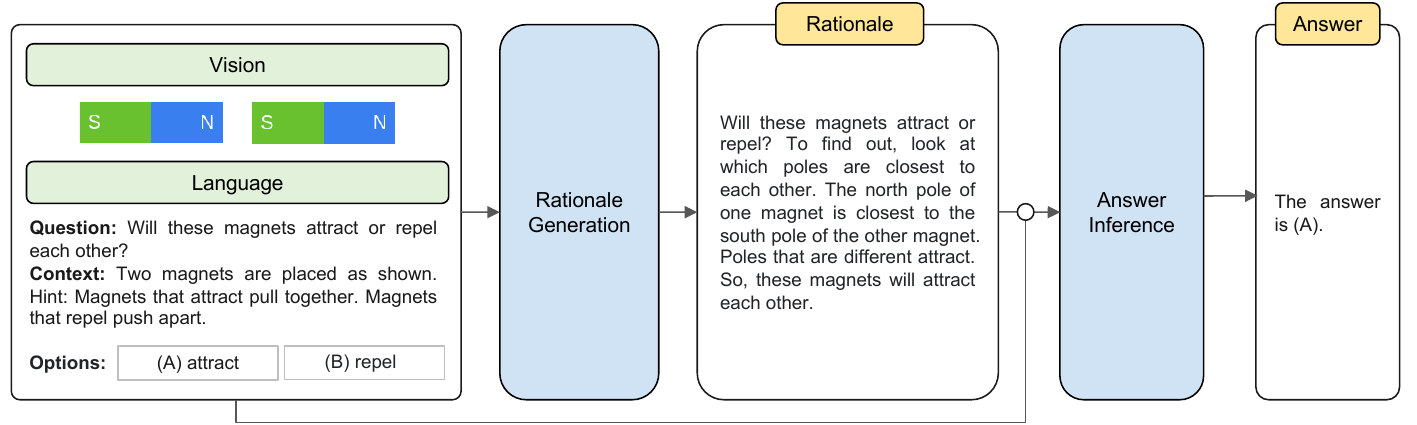}
  \end{center}
  \vspace{-3mm}
  \caption{Overview of our Multimodal-CoT framework. Multimodal-CoT consists of two stages: (i) rationale generation and (ii) answer inference. Both stages share the same model structure but differ in the input and output. In the first stage, we feed the model with language and vision inputs to generate rationales. In the second stage, we append the original language input with the rationale generated from the first stage. Then, we feed the updated language input with the original vision input to the model to infer the answer.}
  \label{fig_overview}
  \vspace{-3mm}
\end{figure*}

\subsection{Framework Overview}
Multimodal-CoT consists of two operation stages: (i) rationale generation and (ii) answer inference. Both stages share the same model structure but differ in the input $X$ and output $Y$. The overall procedure is illustrated in Figure~\ref{fig_overview}. We will take vision-language as an example to show how Multimodal-CoT works.

In the rationale generation stage, we feed the model with $X=\{X_{\textrm{language}}^{1}, X_{\textrm{vision}}\}$ where $X_{\textrm{language}}^{1}$ represents the language input in the first stage and $X_{\textrm{vision}}$ represents the vision input, i.e., the image. For example, $X$ can be instantiated as a concatenation of question, context, and options of a multiple choice reasoning problem \citep{lu2022learn} as shown in Figure \ref{fig_overview}. The goal is to learn a rationale generation model ${R} = F(X)$ where ${R}$ is the rationale.

In the answer inference stage, the rationale $R$ is appended to the original language input $X_{\textrm{language}}^{1}$ to construct the language input in the second stage, $X_{\textrm{language}}^{2} = X_{\textrm{language}}^{1} \circ {R}$ where $\circ$ denotes concatenation. Then, we feed the updated input $X'=\{X_{\textrm{language}}^{2}, X_{\textrm{vision}}\}$ to the answer inference model to infer the final answer $A = F(X')$.

{In both stages, we train two models with the same architecture independently.} They take the annotated elements (e.g., $X\rightarrow R$, $XR\rightarrow A$, respectively) from the training set for supervised learning. During inference, given $X$, the rationales for the test sets are generated using the model trained in the first stage; they are used in the second stage for answer inference.

\subsection{Model Architecture}
Given language input $X_{\textrm{language}} \in \{X_{\textrm{language}}^{1}, X_{\textrm{language}}^{2}\}$ and vision input $X_{\textrm{vision}}$, we compute the probability of generating target text $Y$ (either the rationale or the answer in Figure~\ref{fig_overview}) of length $N$ by
\begin{equation}
	\label{eq:gated}
	p(Y|X_{\textrm{language}},X_{\textrm{vision}}) = \prod_{i=1}^{N} p_{\theta}\left(Y_{i} \mid X_{\textrm{language}}, X_{\textrm{vision}}, Y_{<i}\right),
\end{equation}
where $p_{\theta}\left(Y_{i} \mid X_{\textrm{language}}, X_{\textrm{vision}}, Y_{<i}\right)$ is implemented with a Transformer-based network \citep{vaswani2017attention}. The network has three major procedures: encoding, interaction, and decoding. Specifically, we feed the language text into a Transformer encoder to obtain a textual representation, which is interacted and fused with the vision representation before being fed into the Transformer decoder.

\paragraph{Encoding} The model $F(X)$ takes both the language and vision inputs and obtains the text representation $H_{\textrm{language}}$ and the image feature $H_{\textrm{vision}}$ by the following functions:
\begin{eqnarray}
	H_{\textrm{language}} & = & \textrm{LanguageEncoder}(X_{\textrm{language}}), \\
	H_{\textrm{vision}}  & = & W_{h} \cdot \textrm{VisionExtractor}(X_{\textrm{vision}}),
	\label{eq:extractor}
\end{eqnarray}
where \textrm{LanguageEncoder}($\cdot$) is implemented as a Transformer model. We use the hidden states of the last layer in the Transformer encoder as the language representation $H_{\textrm{language}} \in \mathbb{R}^{n \times d}$ where $n$ denotes the length of the language input, and $d$ is the hidden dimension.
Meanwhile, \textrm{VisionExtractor}($\cdot$) is used to vectorize the input image into vision features. Inspired by the recent success of Vision Transformers \citep{dosovitskiy2020image}, we fetch the patch-level features by {frozen} vision extraction models, such as ViT \citep{dosovitskiy2021image}. After obtaining the patch-level vision features, we apply a learnable projection matrix $W_{h}$ to convert the shape of $\textrm{VisionExtractor}(X_{\textrm{vision}})$ into that of $H_{\textrm{language}}$; thus we have $H_{\textrm{vision}} \in \mathbb{R}^{m \times d}$ where $m$ is the number of patches.

Note that our approach is general to both scenarios with or without image context. For the questions without associated images, we use all-zero vectors as the ``blank features'' with the same shape as the normal image features to tell the model to ignore them.

\paragraph{Interaction} After obtaining language and vision representations, we use a single-head attention network to correlate text tokens with image patches, where the query (${Q}$), key (${K}$) and value (${V}$) are $H_{\textrm{language}}$, $H_{\textrm{vision}}$ and $H_{\textrm{vision}}$, respectively. The attention output $H_{\textrm{vision}}^\textrm{attn} \in \mathbb{R}^{n \times d}$ is defined as:
$
	H_{\textrm{vision}}^\textrm{attn} = \textrm{Softmax}(\frac{{Q}{{K}^{\top}}}{\sqrt{d_k}}){V},
	  \label{eq:selective_attn}
$
where $d_k$ is the same as the dimension of $H_{\textrm{language}}$ because a single head is used. 

Then, we apply the gated fusion mechanism \citep{zhang2020neural,wu2021good,li2022vision} to fuse $H_{\textrm{language}}$ and $H_{\textrm{vision}}$. The fused output $H_{\textrm{fuse}} \in \mathbb{R}^{n \times d}$ is obtained by:
\begin{eqnarray}
	\lambda & = & \textrm{Sigmoid}(W_{l}{H_{\textrm{language}}} + W_{v}{H_{\textrm{vision}}^\textrm{attn}}), \label{eq:gate}\\
	H_{\textrm{fuse}} & = & (1 - \lambda) \cdot H_{\textrm{language}} + \lambda \cdot H_{\textrm{vision}}^\textrm{attn} ,\label{eq:gated_fusion}
\end{eqnarray}
\noindent where $W_{l}$ and $W_{v}$ are learnable parameters. 

\paragraph{Decoding} Finally, the fused output $H_{\textrm{fuse}}$ is fed into the  Transformer decoder to predict the target $Y$. 

\section{Experiments}
This section will present the benchmark dataset, the implementation of our technique, and the baselines for comparisons. Then, we will report our main results and findings.
\subsection{Dataset}
Our method is evaluated on the ScienceQA \citep{lu2022learn} and A-OKVQA \citep{schwenk2022okvqa} benchmark datasets. 
We choose those datasets because they are latest multimodal reasoning benchmarks with annotated reasoning chains. 
ScienceQA is a large-scale multimoda science question dataset with annotated lectures and explanations. It contains 21$k$ multimodal multiple choice questions with rich domain diversity across 3 subjects, 26 topics, 127 categories, and 379 skills. There are 12$k$, 4$k$, and 4$k$ questions in the training, validation, and test splits, respectively. A-OKVQA is a knowledge-based visual question answering benchmark, which has 25$k$ questions requiring a broad base of commonsense and world knowledge to answer. It has 17$k$/1$k$/6$k$ questions for train/val/test. As A-OKVQA provides multiple-choice and direct answer evaluation settings, we use the multiple-choice setting to keep consistency with ScienceQA.

\subsection{Implementation}\label{app-sec:baseline}
The following part presents the experimental settings of Multimodal-CoT and the baseline methods. 
\paragraph{Experimental Settings} We adopt the T5 encoder-decoder architecture \citep{raffel2020exploring} under $\texttt{Base}$ (200M) and $\texttt{large}$ (700M) settings in our framework. We apply FLAN-Alpaca to initialize our model weights.\footnote{\url{https://github.com/declare-lab/flan-alpaca}.} We will show that Multimodal-CoT is generally effective with other backbone LMs, such as UnifiedQA \citep{khashabi2020unifiedqa} and FLAN-T5 \citep{chung2022scaling} (Section \ref{sec:backbones}). The vision features are obtained by the frozen ViT-large encoder \citep{dosovitskiy2021image}. We fine-tune the models up to 20 epochs, with a learning rate of 5e-5. The maximum input sequence length is 512. The batch size is 8. Our experiments are run on 8 NVIDIA Tesla V100 32G GPUs. More details are presented in Appendix \ref{app:exp}.

\begin{table*}[t]
\centering
\vspace{-3mm}
\caption{Main results (\%). Size = backbone model size from the ScienceQA leaderboard (``-'' means unavailable or unknown). Question classes: NAT = natural science, SOC = social science, LAN = language science, TXT = text context, IMG = image context, NO = no context, G1-6 = grades 1-6, G7-12 = grades 7-12. Segment 1: Human performance; Segment 2: VQA baselines; Segment 3: LM baselines, i.e., UnifiedQA and few-shot learning LLMs; Segment 4: Fine-tuned large vision-language models; Segment 5: Our Multimodal-CoT results. Prior published best results are marked with an \underline{underline}. Our best average result is in \textbf{bold} face. $\dagger$ denotes concurrent studies, either through citation or comparison with Multimodal-CoT. }
\small
\renewcommand\tabcolsep{5pt} 
\resizebox{1.0\linewidth}{!}
{
\begin{tabular}{l|rcccccccc|l} 
\toprule
 Model  & Size & NAT & SOC & LAN & TXT & IMG & NO & G1-6 & G7-12 & ~Avg \\
\midrule
Human & - & {90.23}  & 84.97 & 87.48 & {89.60} & {87.50} & \fix{88.10} & 91.59 & 82.42 & 88.40 \\
 \midrule
 MCAN \citep{yu2019mcan}  & 95M& 56.08 & 46.23 & 58.09 & 59.43 & 51.17 & \fix{55.40} & 51.65 & 59.72 & 54.54 \\
 Top-Down \citep{Anderson2017up} & 70M & 59.50 & 54.33 & 61.82 & 62.90 & 54.88 & \fix{59.79} & 57.27 & 62.16 & 59.02 \\
 BAN \citep{Kim2018} & 112M & 60.88 & 46.57 & 66.64 & 62.61 & 52.60 & \fix{{65.51}} & 56.83 & 63.94 & 59.37 \\
 DFAF \citep{gao2019dynamic} & 74M & 64.03 & 48.82 & 63.55 & 65.88 & 54.49 & \fix{64.11} & 57.12 & 67.17 & 60.72 \\
 ViLT \citep{pmlr-v139-kim21k} & 113M & 60.48 & 63.89 & 60.27 & 63.20 & 61.38 & \fix{57.00} & 60.72 & 61.90 & 61.14 \\
 Patch-TRM \citep{lu2021iconqa} & 90M & {65.19} & 46.79 & {65.55} & {66.96} & 55.28 & \fix{64.95} & 58.04 & {67.50} & 61.42 \\
 VisualBERT \citep{li2019visualbert} & 111M & 59.33 & {69.18} & 61.18 & 62.71 & {62.17} & \fix{58.54} & {62.96} & 59.92 & {61.87} \\
  \midrule
 UnifiedQA \citep{lu2022learn} & 223M &{71.00} & {76.04} & {78.91} & {66.42} & {66.53} & \fix{{81.81}} & {77.06} & 68.82 & {74.11}  \\

GPT-3.5 (text-davinci-002) \citep{lu2022learn} & 173B & {75.44} & {70.87} & {78.09} & {74.68} & {67.43} & {79.93} & {78.23} & {69.68} & {75.17} \\
GPT-3.5 (text-davinci-003)& 173B & 77.71 & 68.73 & 80.18 & 75.12 & 67.92 & 81.81 & 80.58 & 69.08 & 76.47 \\
ChatGPT \citep{lu2023chameleon} & - & 78.82 & 70.98 & 83.18 & 77.37 & 67.92 & 86.13 & 80.72 & 74.03 & 78.31	\\
GPT-4  \citep{lu2023chameleon} & - & 85.48 & 72.44 & 90.27 & 82.65 & 71.49 & 92.89 & 86.66 & 79.04 & 83.99\\
Chameleon (ChatGPT) \citep{lu2023chameleon}$\dagger$  & - &  81.62 &  70.64 &  84.00 &  79.77 &  70.80 &  86.62 &  81.86 & 76.53 &  79.93\\
Chameleon (GPT-4) \citep{lu2023chameleon}$\dagger$ & - & 89.83 & 74.13	 & {89.82} & 88.27 & 77.64 & 92.13 & 88.03 & 83.72 &  \underline{86.54} \\
\midrule  
LLaMA-Adapter \citep{zhang2023llama}$\dagger$ & 6B & 84.37  & 88.30  & 84.36  & 83.72  & 80.32  & 86.90  & 85.83 &  84.05  & 85.19 \\
LLaVA \citep{liu2023visual}$\dagger$  & 13B & 90.36  & 95.95  & 88.00  & 89.49  & 88.00 &  90.66  & 90.93  & 90.90  & 90.92 \\
InstructBLIP \citep{instructblip}$\dagger$  & 11B& - & - & - & - & 90.70 & - & - & - & \\
 \midrule
 Mutimodal-CoT$_\texttt{Base}$  & 223M & 84.06 & 92.35 & 82.18 & 82.75 & 82.75 & 84.74 & 85.79 & 84.44 & 85.31 \\
 Mutimodal-CoT$_\texttt{Large}$ & 738M & 91.03 & {93.70} & 86.64 & 90.13 & 88.25 & 89.48 & {91.12} & {89.26} & {90.45} \\
 \bottomrule
\end{tabular}
}
 \label{tab:main_results}
\end{table*}

\paragraph{Baseline Models}
We utilized three categories of methods as our baselines:

(i) Visual question answering (VQA) models, including MCAN \citep{yu2019mcan}, Top-Down \citep{Anderson2017up}, BAN \citep{Kim2018}, DFAF \citep{gao2019dynamic}, ViLT \citep{pmlr-v139-kim21k}, Patch-TRM \citep{lu2021iconqa}, and VisualBERT \citep{li2019visualbert}. These VQA baselines take the question, context, and choices as textual input, while utilizing the image as visual input. They employ a linear classifier to predict the score distribution over the choice candidates.

(ii) LMs, including the text-to-text UnifiedQA model \citep{khashabi2020unifiedqa} and few-shot learning LLMs (GPT-3.5, ChatGPT, GPT-4, and Chameleon \citep{lu2023chameleon}). UnifiedQA \citep{khashabi2020unifiedqa} is adopted as it is the best fine-tuning model in \citet{lu2022learn}. UnifiedQA takes the textual information as the input and outputs the answer choice. The image is converted into a caption extracted by an image captioning
model following \citet{lu2022learn}. UnifiedQA treats our task as a text generation problem. In \citet{lu2022learn}, it is trained to generate a target answer text, i.e., one of the candidate options. Then, the most similar option is selected as the final prediction to evaluate the question answering accuracy. For GPT-3.5 models \citep{chen2020big}, we use the text-davinci-002 and text-davinci-003 engines due to their strong performance. In addition, we also include the comparison with ChatGPT and GPT-4. The inference is based on the few-shot prompting, where two in-context examples from the training set are concatenated before the test instance. The few-shot demonstrations are the same as those in \citet{lu2022learn}.

(iii) Fine-tuned large vision-language model. We select the recently released LLaMA-Adapter \citep{zhang2023llama}, LLaVA \citep{liu2023visual}, and InstructBLIP \citep{instructblip} as the competitive large vision-language baselines. For LLaMA-Adapter, the backbone model is the 7B LLaMA model fine-tuned with 52$k$ self-instruct demonstrations. To adapt to our tasks, the model is further fine-tuned on the ScienceQA dataset.


\subsection{Main Results}
Table \ref{tab:main_results} shows the main results in the ScienceQA benchmark. We observe that Mutimodal-CoT$_\texttt{Large}$ achieves substantial performance gains over the prior best model in publications (86.54\%$\rightarrow$90.45\%). 
The efficacy of Multimodal-CoT is further supported by the results obtained from the A-OKVQA benchmark in Table \ref{tab:vqa-results}.  

\begin{wraptable}{r}{0.48\textwidth}
\centering
\vspace{-7.8mm}
\caption{Results on A-OKVQA. Baseline results are from \citep{chen2023see} and \citet{schwenk2022okvqa}.}
         \setlength{\tabcolsep}{12pt}
         \small       
{
\begin{tabular}{lr}
\toprule
 Model  & Accuracy \\
 \midrule
 BERT & 32.93 \\
 GPT-3 (Curie) & 35.07 \\
 \midrule
 IPVR (OPT-66B)  & 48.6 \\
 ViLBERT & 49.1 \\
\midrule
Language-only Baseline & 47.86 \\
Multimodal-CoT$_\texttt{Base}$ & 50.57 \\
 \bottomrule
\end{tabular}
}
\vspace{-7.8mm}
 \label{tab:vqa-results}
\end{wraptable}


It is worth noting that Chameleon, LLaMA-Adapter, LLaVA, and InstructBLIP are concurrent works released several months after our work. In the subsequent Section \ref{section:large}, we will show that our method is orthogonal to those multimodal models (e.g., InstructBLIP) and can be potentially used with them together to improve generality further, i.e., scaled to scenarios where human-annotated rationales are unavailable, thereby establishing the effectiveness across diverse tasks.

Ablation study results in Table \ref{tab:abl_results} show that both the integration of vision features and the two-stage framework design contribute to the overall performance.
\begin{table}[htb]
\centering
\caption{Ablation results of Multimodal-CoT.}
{
\begin{tabular}{lcc}
\toprule
 Model  & Base  & Large \\
 \midrule
Multimodal-CoT  & 85.31 & 90.45 \\
\quad w/o Two-Stage Framework & 82.62 &  84.56\\
\quad w/o Vision Features & 78.57 &  83.97 \\
 \bottomrule
\end{tabular}
}
 \label{tab:abl_results}
\end{table}

Furthermore, we find that Multimodal-CoT demonstrates the ability to mitigate hallucination (Section \ref{sec:multimodal}) and improve convergence (Section \ref{appendix:cover}).


\section{Analysis}
The following analysis will first show that Multimodal-CoT helps enhance convergence speed and has the feasibility of adaptation to scenarios without human-annotated rationales. Then, we investigate the general effectiveness of Multimodal-CoT with different backbone models and vision features. We will also conduct an error analysis to explore the limitations to inspire future studies. We use models under the $\texttt{base}$ size for analysis unless otherwise stated.

\begin{wrapfigure}{r}{0.536\textwidth}
	\centering
 \vspace{-4mm}
\pgfplotsset{height=5.1cm,width=9.35cm,compat=1.14,every axis/.append style={thick},every tick label/.append style={font=\small},every axis legend/.append style={ at={(0.5,0.95)}},legend columns=2 row=2} 
\begin{tikzpicture} \tikzset{every node}=[font=\small] 
\begin{axis} [
                align = center,
                legend style={at={(0.5,1.3)},anchor=north},
                ymin=52, ymax=90,
                xticklabels={1,2,3,4,5,6,7,8,9,10}, xtick={0,1,2,3,4,5,6,7,8,9},
                ylabel style={align=center},
                xlabel={Epoch},
                ylabel={Accuracy},
                ytick={50, 60, 70, 80, 90},
                xtick pos=bottom,
                ytick pos=left,
                ]

\addplot+[color=brown,
                    mark=triangle,
                    mark size=1.5pt,
                    ] coordinates { (0, 57.62)  (1, 66.49) (2, 73.14) (3, 77.67) (4, 78.94) (5, 80.05) (6, 80.97) (7, 81.60) (8, 81.29) (9, 81.63)};
\addlegendentry{\small {One-stage} Baseline}

\addplot+ [color=brown,
                    mark=star,
                    mark size=1.5pt,
                    ] coordinates {(0, 65.73) (1, 75.64) (2, 78.84) (3, 80.05) (4, 80.75) (5, 81.33) (6, 81.51) (7, 82.07) (8, 81.91) (9, 82.38)};
\addlegendentry{\small {One-stage} Multimodal}

\addplot+ [color=blue,
                    mark=square,
                    mark size=2.5pt,
                    ] coordinates { (0, 77.55)  (1, 78.75) (2, 78.61) (3, 78.89) (4, 79.15) (5, 78.92) (6, 78.87) (7, 79.13) (8, 79.25) (9, 78.57)};
\addlegendentry{\small Two-Stage Baseline}

\addplot+ [color=blue,
                    mark=o,
                    mark size=2.5pt,
                    ] coordinates { (0,  81.27)  (1, 81.82) (2,  83.47) (3,  83.40) (4,  84.08) (5,  84.41) (6,  84.43) (7,  84.27) (8, 84.50) (9,  85.31)};
\addlegendentry{\small Two-Stage Multimodal}

\end{axis}
\end{tikzpicture}
\vspace{-6.8mm}
	\caption{Accuracy curve of the {No-CoT baseline} and Multimodal-CoT variants.}\label{fig:curve}
 \vspace{-3mm}
\end{wrapfigure}
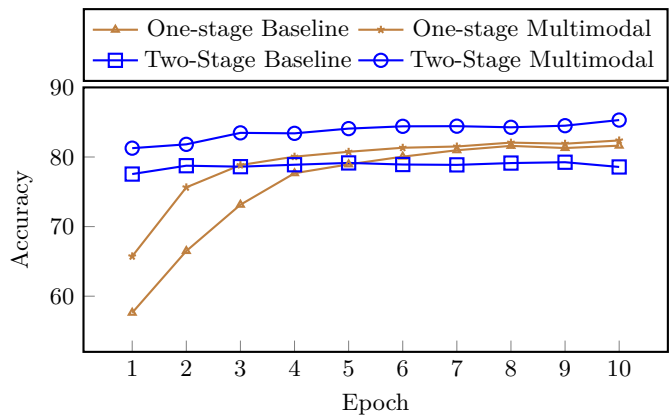
\subsection{Multimodality Boosts Convergence}\label{appendix:cover}
Figure \ref{fig:curve} shows the validation accuracy curve of the baseline and Multimodal-CoT across different training epochs. {``One-stage'' is based on the QCM$\rightarrow$A input-output format as it achieves the best performance in Table 2 and ``Two-stage'' is our two-stage framework.} We find that the two-stage methods achieve relatively higher accuracy at the beginning than the {one-stage} baselines that generate the answer directly without CoT. However, without the vision features, the two-stage baseline could not yield better results as the training goes on due to low-quality rationales (as observed in Section \ref{sec:prelim}). In contrast, 
using vision features helps generate more effective rationales that contribute to better answer accuracy in our two-stage multimodal variant.

\subsection{When Multimodal-CoT Meets Large Models}\label{section:large}
A recent flame is to leverage large language models or large vision-language models to generate reasoning chains for multimodal question answering problems \citep{zhang2023llama,lu2023chameleon,liu2023visual,alayrac2022flamingo,hao2022language,yasunaga2022retrieval}. We are interested in whether we can use large models to generate the rationales for Multimodal-CoT; thus breaking the need for datasets with human-annotated rationales. During the first-stage training of Multimodal-CoT, our target rationales are based on human annotation in the benchmark datasets. Now, we replace the target rationales with those generated ones. {As ScienceQA contains questions with images and without images, we leverage InstructBLIP and ChatGPT to generate the rationales for questions with paired images and questions without paired images, respectively.\footnote{Examples are provided in Appendix \ref{appendix:large}.} Then, we combine both of the generated pseudo-rationales as the target rationales for training (Multimodal-CoT w/ Generation) instead of relying on the human annotation of reasoning chains (Multimodal-CoT w/ Annotation).}


\begin{table}[htb]
\centering
\caption{Result comparison with large models. We also present the results of InstructBLIP and ChatGPT baselines for reference. The inference format for the two baselines is QCM$\rightarrow$A.}
{
\begin{tabular}{lccc}
\toprule
 Model  & IMG & TXT & AVG \\
 \midrule
 InstructBLIP & 60.50 & - & -\\
 ChatGPT & 56.52 & 67.16  & 65.95 \\
\midrule
Multimodal-CoT w/ Annotation & 88.25 & 90.13  & 90.45 \\
Multimodal-CoT w/ Generation & 83.54 & 85.73& 87.76\\ 
 \bottomrule
\end{tabular}
}
 \label{tab:rationale_gen}
\end{table}

Table \ref{tab:rationale_gen} shows the comparison results. We see that using the generated rationales achieves comparable performance to using human-annotated rationales for training. In addition, the performance is also much better than directly prompting those baseline models to obtain the answer (in the QCM$\rightarrow$A inference format).

We see that Multimodal-CoT can work effectively with large models. The findings above compellingly show the feasibility of adaptation to scenarios without human-annotated rationales, thereby establishing the effectiveness of our approach across diverse tasks.


\subsection{Effectiveness Across Backbones}\label{sec:backbones}
To test the generality of the benefits of our approach to other backbone models, we alter the underlying LMs to other variants in different types. As shown in Table \ref{tab:backbone}, our approach is generally effective for the widely used backbone models. 

\begin{minipage}{\textwidth}
        \begin{minipage}[t]{0.48\textwidth}
            \centering
            \makeatletter\def\@captype{table}\makeatother\caption{Using different backbone LMs. \label{tab:backbone}}
            \setlength{\tabcolsep}{8pt}
            \begin{tabular}{lc}\toprule
 {Method}   & {Accuracy} \\
 \midrule
 Prior Best \citep{lu2022learn} & 75.17 \\
 \midrule
 MM-CoT on UnifiedQA & 82.55 \\
 MM-CoT on FLAN-T5  & 83.19 \\
 MM-CoT on FLAN-Alpaca &  85.31 \\
\bottomrule
\end{tabular}
        \end{minipage}
        \begin{minipage}[t]{0.48\textwidth}
        \centering
        \makeatletter\def\@captype{table}\makeatother
        \caption{Using different vision features. \label{tab:visual_features}}
        \setlength{\tabcolsep}{10pt}
            \begin{tabular}{lcc}\toprule
 {Feature}  & {Feature Shape}& {Accuracy}\\\midrule
 ViT & (145, 1024) & 85.31 \\
 CLIP  & (49, 2048) & 84.27 \\
 DETR   &  (100, 256) & 83.16\\
 ResNet & (512, 2048) & 82.86 \\

\bottomrule
\end{tabular}
        \end{minipage}
    \end{minipage}
    


\subsection{Using Different Vision Features}\label{sec:vision_features}
Different vision features may affect the model performance. We compare three widely-used types of vision features, ViT \citep{dosovitskiy2021image}, CLIP \citep{radford2021learning}, DETR \citep{carion2020end}, and ResNet \citep{he2016deep}. ViT, CLIP, and DETR are patch-like features. For the ResNet features, we repeat the pooled features of ResNet-50 to the same length with the text sequence to imitate the patch-like features, where each patch is the same as the pooled image features. More details of the vision features are presented in Appendix \ref{appendix:vision_features}.

Table \ref{tab:visual_features} shows the comparative results of vision features. We observe that ViT achieves {relatively better} performance. Therefore, we use ViT by default in Multimodal-CoT.



\subsection{Alignment Strategies for Multimodal Interaction}\label{appendix:aligh}

We are interested in whether using different alignment strategies for multimodal interaction may contribute to different behaviors of multimodal-CoT. To this end, we tried another alignment strategy, i.e., image-grounded text encoder, in BLIP \cite{li2022blip}. This alignment approach injects visual information by inserting one additional cross-attention layer between the self-attention layer and the feed-forward network for each transformer block of the text encoder. Our current strategy in the paper is similar to the unimodal encoder as in BLIP, which is used for comparison. In Table \ref{tab:alignment}, we see that using other alignment strategies also contributes to better performance than direct answering.

\begin{table}[htb]
\centering
\caption{Result comparison with different alignment strategies for multimodal interaction.}
{
\begin{tabular}{lc}
\toprule
 Model  & Accuracy \\
 \midrule
 Direct Answering &  82.62 \\
 Unimodal encoder & 	85.31 \\
 Image-grounded text encoder & 84.60 \\
 \bottomrule
\end{tabular}
}
 \label{tab:alignment}
\end{table}

\subsection{{Generalization to Other Multimodal Reasoning Benchmarks}}\label{appendix:aligh}
{We are interested in evaluating the generalization capability of Multimodal-CoT to datasets outside its training domain. For this purpose, we utilize the widely-recognized multimodal reasoning benchmark, MMMU \citep{yue2023mmmu}, and conduct an evaluation of Multimodal-CoT on MMMU without further training.}

\begin{table}[htb]
\centering
\caption{{Generalization performance on MMMU.}}
{
\begin{tabular}{lrc}
\toprule
 Model  & Size & Accuracy \\
 \midrule
 Kosmos-2 \citep{peng2024grounding} & 1.6B & 24.4 \\
 Fuyu \citep{bavishi2024fuyu} & 8B & 27.9 \\
 OpenFlamingo-2 \citep{awadalla2023openflamingo} & 9B &  28.7\\ 
 MiniGPT4-Vicuna \citep{zhu2023minigpt} & 13B & 26.8 \\ 
 Multimodal-CoT & 738M & 28.7 \\
 \midrule
 GPT-4V(ision) \citep{gpt4v} & - & 56.8 \\
 Gemini Ultra \citep{reid2024gemini} & - & 59.4 \\
 \bottomrule
\end{tabular}
}
 \label{tab:transfer}
\end{table}

{As shown in Table \ref{tab:transfer}, it is evident that Multimodal-CoT demonstrates effective generalization to MMMU, achieving better performance than various larger models around 8B.}

\subsection{Error Analysis}\label{sec:case_studies}

\begin{wrapfigure}{r}{0.45\textwidth}
  \begin{center}
    \vspace{-5mm}
   \includegraphics[width=0.38\textwidth]{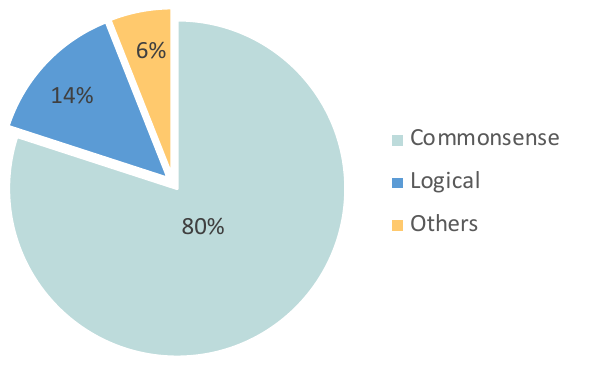}
  \end{center}
  \vspace{-6mm}
  \caption{Categorization analysis.}
    \vspace{-3mm}
  \label{fig:analysis_error}
\end{wrapfigure}

To gain deeper insights into the behavior of Multimodal-CoT and facilitate future research, we manually analyzed randomly selected examples generated by our approach. The categorization results are illustrated in Figure \ref{fig:analysis_error}. We examined 50 samples that yielded incorrect answers and categorized them accordingly. The examples from each category can be found in Appendix \ref{appendix:case_study}.

The most prevalent error type is commonsense mistakes, accounting for 80\% of the errors. These mistakes occur when the model is faced with questions that require commonsense knowledge, such as interpreting maps, counting objects in images, or utilizing the alphabet. The second error type is logical mistakes, constituting 14\% of the errors, which involve contradictions in the reasoning process. Additionally, we have observed cases where incorrect answers are provided despite the CoT being either empty or correct, amounting to 6\% of the errors. The CoT in these cases may not necessarily influence the final answer.

The analysis reveals potential avenues for future research. Enhancements can be made to Multimodal-CoT by: (i) integrating more informative visual features and strengthening the interaction between language and vision to enable comprehension of maps and numerical counting; (ii) incorporating commonsense knowledge; and (iii) implementing a filtering mechanism, such as using only relevant CoTs to infer answers and disregarding irrelevant ones.

\section{Conclusion}
This paper formally studies the problem of multimodal CoT. {We propose Multimodal-CoT that incorporates language and vision modalities into a two-stage framework that separates rationale generation and answer inference, so answer inference can leverage better generated rationales from multimodal information.} With Multimodal-CoT, our model under 1 billion parameters achieves state-of-the-art performance on the ScienceQA benchmark. Analysis shows that Multimodal-CoT has the merits of mitigating hallucination and enhancing convergence speed. Our error analysis identifies the potential to leverage more effective vision features, inject commonsense knowledge, and apply filtering mechanisms to improve CoT reasoning in future studies.

\bibliography{main}

\begin{thebibliography}{65}
\providecommand{\natexlab}[1]{#1}
\providecommand{\url}[1]{\texttt{#1}}
\expandafter\ifx\csname urlstyle\endcsname\relax
  \providecommand{\doi}[1]{doi: #1}\else
  \providecommand{\doi}{doi: \begingroup \urlstyle{rm}\Url}\fi

\bibitem[Alayrac et~al.(2022)Alayrac, Donahue, Luc, Miech, Barr, Hasson, Lenc, Mensch, Millican, Reynolds, et~al.]{alayrac2022flamingo}
Jean-Baptiste Alayrac, Jeff Donahue, Pauline Luc, Antoine Miech, Iain Barr, Yana Hasson, Karel Lenc, Arthur Mensch, Katherine Millican, Malcolm Reynolds, et~al.
\newblock Flamingo: a visual language model for few-shot learning.
\newblock \emph{Advances in Neural Information Processing Systems}, 35:\penalty0 23716--23736, 2022.

\bibitem[Anderson et~al.(2018)Anderson, He, Buehler, Teney, Johnson, Gould, and Zhang]{Anderson2017up}
Peter Anderson, Xiaodong He, Chris Buehler, Damien Teney, Mark Johnson, Stephen Gould, and Lei Zhang.
\newblock Bottom-up and top-down attention for image captioning and visual question answering.
\newblock In \emph{2018 {IEEE} Conference on Computer Vision and Pattern Recognition, {CVPR} 2018, Salt Lake City, UT, USA, June 18-22, 2018}, pp.\  6077--6086. {IEEE} Computer Society, 2018.
\newblock \doi{10.1109/CVPR.2018.00636}.

\bibitem[Awadalla et~al.(2023)Awadalla, Gao, Gardner, Hessel, Hanafy, Zhu, Marathe, Bitton, Gadre, Sagawa, et~al.]{awadalla2023openflamingo}
Anas Awadalla, Irena Gao, Josh Gardner, Jack Hessel, Yusuf Hanafy, Wanrong Zhu, Kalyani Marathe, Yonatan Bitton, Samir Gadre, Shiori Sagawa, et~al.
\newblock Openflamingo: An open-source framework for training large autoregressive vision-language models.
\newblock \emph{arXiv preprint arXiv:2308.01390}, 2023.

\bibitem[Bavishi et~al.(2024)Bavishi, Elsen, Hawthorne, Nye, Odena, Somani, and Ta{\c{s}}{\i}rlar]{bavishi2024fuyu}
Rohan Bavishi, Erich Elsen, Curtis Hawthorne, Maxwell Nye, Augustus Odena, Arushi Somani, and Sa{\u{g}}nak Ta{\c{s}}{\i}rlar.
\newblock Fuyu-8b: A multimodal architecture for ai agents, 2024.

\bibitem[Brown et~al.(2020)Brown, Mann, Ryder, Subbiah, Kaplan, Dhariwal, Neelakantan, Shyam, Sastry, Askell, Agarwal, Herbert{-}Voss, Krueger, Henighan, Child, Ramesh, Ziegler, Wu, Winter, Hesse, Chen, Sigler, Litwin, Gray, Chess, Clark, Berner, McCandlish, Radford, Sutskever, and Amodei]{brown2020language}
Tom~B. Brown, Benjamin Mann, Nick Ryder, Melanie Subbiah, Jared Kaplan, Prafulla Dhariwal, Arvind Neelakantan, Pranav Shyam, Girish Sastry, Amanda Askell, Sandhini Agarwal, Ariel Herbert{-}Voss, Gretchen Krueger, Tom Henighan, Rewon Child, Aditya Ramesh, Daniel~M. Ziegler, Jeffrey Wu, Clemens Winter, Christopher Hesse, Mark Chen, Eric Sigler, Mateusz Litwin, Scott Gray, Benjamin Chess, Jack Clark, Christopher Berner, Sam McCandlish, Alec Radford, Ilya Sutskever, and Dario Amodei.
\newblock Language models are few-shot learners.
\newblock In Hugo Larochelle, Marc'Aurelio Ranzato, Raia Hadsell, Maria{-}Florina Balcan, and Hsuan{-}Tien Lin (eds.), \emph{Advances in Neural Information Processing Systems 33: Annual Conference on Neural Information Processing Systems 2020, NeurIPS 2020, December 6-12, 2020, virtual}, 2020.

\bibitem[Carion et~al.(2020)Carion, Massa, Synnaeve, Usunier, Kirillov, and Zagoruyko]{carion2020end}
Nicolas Carion, Francisco Massa, Gabriel Synnaeve, Nicolas Usunier, Alexander Kirillov, and Sergey Zagoruyko.
\newblock End-to-end object detection with transformers.
\newblock In \emph{Computer Vision--ECCV 2020: 16th European Conference, Glasgow, UK, August 23--28, 2020, Proceedings, Part I}, pp.\  213--229, 2020.

\bibitem[Chen et~al.(2020)Chen, Kornblith, Swersky, Norouzi, and Hinton]{chen2020big}
Ting Chen, Simon Kornblith, Kevin Swersky, Mohammad Norouzi, and Geoffrey~E. Hinton.
\newblock Big self-supervised models are strong semi-supervised learners.
\newblock In Hugo Larochelle, Marc'Aurelio Ranzato, Raia Hadsell, Maria{-}Florina Balcan, and Hsuan{-}Tien Lin (eds.), \emph{Advances in Neural Information Processing Systems 33: Annual Conference on Neural Information Processing Systems 2020, NeurIPS 2020, December 6-12, 2020, virtual}, 2020.

\bibitem[Chen et~al.(2022)Chen, Ma, Wang, and Cohen]{chen2022program}
Wenhu Chen, Xueguang Ma, Xinyi Wang, and William~W Cohen.
\newblock Program of thoughts prompting: Disentangling computation from reasoning for numerical reasoning tasks.
\newblock \emph{ArXiv preprint}, abs/2211.12588, 2022.

\bibitem[Chen et~al.(2023)Chen, Zhou, Shen, Hong, Zhang, and Gan]{chen2023see}
Zhenfang Chen, Qinhong Zhou, Yikang Shen, Yining Hong, Hao Zhang, and Chuang Gan.
\newblock See, think, confirm: Interactive prompting between vision and language models for knowledge-based visual reasoning.
\newblock \emph{ArXiv preprint}, abs/2301.05226, 2023.

\bibitem[Chowdhery et~al.(2022)Chowdhery, Narang, Devlin, Bosma, Mishra, Roberts, Barham, Chung, Sutton, Gehrmann, Schuh, Shi, Tsvyashchenko, Maynez, Rao, Barnes, Tay, Shazeer, Prabhakaran, Reif, Du, Hutchinson, Pope, Bradbury, Austin, Isard, Gur-Ari, Yin, Duke, Levskaya, Ghemawat, Dev, Michalewski, Garcia, Misra, Robinson, Fedus, Zhou, Ippolito, Luan, Lim, Zoph, Spiridonov, Sepassi, Dohan, Agrawal, Omernick, Dai, Pillai, Pellat, Lewkowycz, Moreira, Child, Polozov, Lee, Zhou, Wang, Saeta, Diaz, Firat, Catasta, Wei, Meier-Hellstern, Eck, Dean, Petrov, and Fiedel]{palm}
Aakanksha Chowdhery, Sharan Narang, Jacob Devlin, Maarten Bosma, Gaurav Mishra, Adam Roberts, Paul Barham, Hyung~Won Chung, Charles Sutton, Sebastian Gehrmann, Parker Schuh, Kensen Shi, Sasha Tsvyashchenko, Joshua Maynez, Abhishek Rao, Parker Barnes, Yi~Tay, Noam Shazeer, Vinodkumar Prabhakaran, Emily Reif, Nan Du, Ben Hutchinson, Reiner Pope, James Bradbury, Jacob Austin, Michael Isard, Guy Gur-Ari, Pengcheng Yin, Toju Duke, Anselm Levskaya, Sanjay Ghemawat, Sunipa Dev, Henryk Michalewski, Xavier Garcia, Vedant Misra, Kevin Robinson, Liam Fedus, Denny Zhou, Daphne Ippolito, David Luan, Hyeontaek Lim, Barret Zoph, Alexander Spiridonov, Ryan Sepassi, David Dohan, Shivani Agrawal, Mark Omernick, Andrew~M. Dai, Thanumalayan~Sankaranarayana Pillai, Marie Pellat, Aitor Lewkowycz, Erica Moreira, Rewon Child, Oleksandr Polozov, Katherine Lee, Zongwei Zhou, Xuezhi Wang, Brennan Saeta, Mark Diaz, Orhan Firat, Michele Catasta, Jason Wei, Kathy Meier-Hellstern, Douglas Eck, Jeff Dean, Slav Petrov, and Noah Fiedel.
\newblock Palm: Scaling language modeling with pathways.
\newblock \emph{ArXiv preprint}, abs/2204.02311, 2022.

\bibitem[Chung et~al.(2022)Chung, Hou, Longpre, Zoph, Tay, Fedus, Li, Wang, Dehghani, Brahma, et~al.]{chung2022scaling}
Hyung~Won Chung, Le~Hou, Shayne Longpre, Barret Zoph, Yi~Tay, William Fedus, Eric Li, Xuezhi Wang, Mostafa Dehghani, Siddhartha Brahma, et~al.
\newblock Scaling instruction-finetuned language models.
\newblock \emph{ArXiv preprint}, abs/2210.11416, 2022.

\bibitem[Dai et~al.(2023)Dai, Li, Li, Tiong, Zhao, Wang, Li, Fung, and Hoi]{instructblip}
Wenliang Dai, Junnan Li, Dongxu Li, Anthony Meng~Huat Tiong, Junqi Zhao, Weisheng Wang, Boyang Li, Pascale Fung, and Steven Hoi.
\newblock Instructblip: Towards general-purpose vision-language models with instruction tuning, 2023.

\bibitem[Dosovitskiy et~al.(2021{\natexlab{a}})Dosovitskiy, Beyer, Kolesnikov, Weissenborn, Zhai, Unterthiner, Dehghani, Minderer, Heigold, Gelly, Uszkoreit, and Houlsby]{dosovitskiy2020image}
Alexey Dosovitskiy, Lucas Beyer, Alexander Kolesnikov, Dirk Weissenborn, Xiaohua Zhai, Thomas Unterthiner, Mostafa Dehghani, Matthias Minderer, Georg Heigold, Sylvain Gelly, Jakob Uszkoreit, and Neil Houlsby.
\newblock An image is worth 16x16 words: Transformers for image recognition at scale.
\newblock In \emph{9th International Conference on Learning Representations, {ICLR} 2021, Virtual Event, Austria, May 3-7, 2021}. OpenReview.net, 2021{\natexlab{a}}.

\bibitem[Dosovitskiy et~al.(2021{\natexlab{b}})Dosovitskiy, Beyer, Kolesnikov, Weissenborn, Zhai, Unterthiner, Dehghani, Minderer, Heigold, Gelly, Uszkoreit, and Houlsby]{dosovitskiy2021image}
Alexey Dosovitskiy, Lucas Beyer, Alexander Kolesnikov, Dirk Weissenborn, Xiaohua Zhai, Thomas Unterthiner, Mostafa Dehghani, Matthias Minderer, Georg Heigold, Sylvain Gelly, Jakob Uszkoreit, and Neil Houlsby.
\newblock An image is worth 16x16 words: Transformers for image recognition at scale.
\newblock In \emph{9th International Conference on Learning Representations, {ICLR} 2021, Virtual Event, Austria, May 3-7, 2021}. OpenReview.net, 2021{\natexlab{b}}.

\bibitem[Fu et~al.(2022)Fu, Peng, Sabharwal, Clark, and Khot]{fu2022complexity}
Yao Fu, Hao Peng, Ashish Sabharwal, Peter Clark, and Tushar Khot.
\newblock Complexity-based prompting for multi-step reasoning.
\newblock \emph{ArXiv preprint}, abs/2210.00720, 2022.

\bibitem[Gao et~al.(2019)Gao, Jiang, You, Lu, Hoi, Wang, and Li]{gao2019dynamic}
Peng Gao, Zhengkai Jiang, Haoxuan You, Pan Lu, Steven C.~H. Hoi, Xiaogang Wang, and Hongsheng Li.
\newblock Dynamic fusion with intra- and inter-modality attention flow for visual question answering.
\newblock In \emph{{IEEE} Conference on Computer Vision and Pattern Recognition, {CVPR} 2019, Long Beach, CA, USA, June 16-20, 2019}, pp.\  6639--6648. Computer Vision Foundation / {IEEE}, 2019.
\newblock \doi{10.1109/CVPR.2019.00680}.

\bibitem[Hao et~al.(2022)Hao, Song, Dong, Huang, Chi, Wang, Ma, and Wei]{hao2022language}
Yaru Hao, Haoyu Song, Li~Dong, Shaohan Huang, Zewen Chi, Wenhui Wang, Shuming Ma, and Furu Wei.
\newblock Language models are general-purpose interfaces.
\newblock \emph{ArXiv preprint}, abs/2206.06336, 2022.

\bibitem[He et~al.(2016)He, Zhang, Ren, and Sun]{he2016deep}
Kaiming He, Xiangyu Zhang, Shaoqing Ren, and Jian Sun.
\newblock Deep residual learning for image recognition.
\newblock In \emph{2016 {IEEE} Conference on Computer Vision and Pattern Recognition, {CVPR} 2016, Las Vegas, NV, USA, June 27-30, 2016}, pp.\  770--778. {IEEE} Computer Society, 2016.
\newblock \doi{10.1109/CVPR.2016.90}.

\bibitem[Ho et~al.(2022)Ho, Schmid, and Yun]{ho2022large}
Namgyu Ho, Laura Schmid, and Se-Young Yun.
\newblock Large language models are reasoning teachers.
\newblock \emph{ArXiv preprint}, abs/2212.10071, 2022.

\bibitem[Huang \& Chang(2022)Huang and Chang]{huang2022towards}
Jie Huang and Kevin Chen-Chuan Chang.
\newblock Towards reasoning in large language models: A survey.
\newblock \emph{ArXiv preprint}, abs/2212.10403, 2022.

\bibitem[Ji et~al.(2022)Ji, Lee, Frieske, Yu, Su, Xu, Ishii, Bang, Madotto, and Fung]{ji2022survey}
Ziwei Ji, Nayeon Lee, Rita Frieske, Tiezheng Yu, Dan Su, Yan Xu, Etsuko Ishii, Yejin Bang, Andrea Madotto, and Pascale Fung.
\newblock Survey of hallucination in natural language generation.
\newblock \emph{ACM Computing Surveys}, 2022.

\bibitem[Khashabi et~al.(2020)Khashabi, Min, Khot, Sabharwal, Tafjord, Clark, and Hajishirzi]{khashabi2020unifiedqa}
Daniel Khashabi, Sewon Min, Tushar Khot, Ashish Sabharwal, Oyvind Tafjord, Peter Clark, and Hannaneh Hajishirzi.
\newblock {UNIFIEDQA}: Crossing format boundaries with a single {QA} system.
\newblock In \emph{Findings of the Association for Computational Linguistics: EMNLP 2020}, pp.\  1896--1907, Online, 2020. Association for Computational Linguistics.
\newblock \doi{10.18653/v1/2020.findings-emnlp.171}.

\bibitem[Khot et~al.(2022)Khot, Trivedi, Finlayson, Fu, Richardson, Clark, and Sabharwal]{khot2022decomposed}
Tushar Khot, Harsh Trivedi, Matthew Finlayson, Yao Fu, Kyle Richardson, Peter Clark, and Ashish Sabharwal.
\newblock Decomposed prompting: A modular approach for solving complex tasks.
\newblock \emph{ArXiv preprint}, abs/2210.02406, 2022.

\bibitem[Kim et~al.(2018)Kim, Jun, and Zhang]{Kim2018}
Jin{-}Hwa Kim, Jaehyun Jun, and Byoung{-}Tak Zhang.
\newblock Bilinear attention networks.
\newblock In Samy Bengio, Hanna~M. Wallach, Hugo Larochelle, Kristen Grauman, Nicol{\`{o}} Cesa{-}Bianchi, and Roman Garnett (eds.), \emph{Advances in Neural Information Processing Systems 31: Annual Conference on Neural Information Processing Systems 2018, NeurIPS 2018, December 3-8, 2018, Montr{\'{e}}al, Canada}, pp.\  1571--1581, 2018.

\bibitem[Kim et~al.(2021)Kim, Son, and Kim]{pmlr-v139-kim21k}
Wonjae Kim, Bokyung Son, and Ildoo Kim.
\newblock Vilt: Vision-and-language transformer without convolution or region supervision.
\newblock In Marina Meila and Tong Zhang (eds.), \emph{Proceedings of the 38th International Conference on Machine Learning, {ICML} 2021, 18-24 July 2021, Virtual Event}, volume 139 of \emph{Proceedings of Machine Learning Research}, pp.\  5583--5594. {PMLR}, 2021.

\bibitem[Kojima et~al.(2022)Kojima, Gu, Reid, Matsuo, and Iwasawa]{kojima2022large}
Takeshi Kojima, Shixiang~Shane Gu, Machel Reid, Yutaka Matsuo, and Yusuke Iwasawa.
\newblock Large language models are zero-shot reasoners.
\newblock \emph{ArXiv preprint}, abs/2205.11916, 2022.

\bibitem[Li et~al.(2022{\natexlab{a}})Li, Lv, Zhou, Zhou, Xiao, Ma, and Zhu]{li2022vision}
Bei Li, Chuanhao Lv, Zefan Zhou, Tao Zhou, Tong Xiao, Anxiang Ma, and JingBo Zhu.
\newblock On vision features in multimodal machine translation.
\newblock In \emph{Proceedings of the 60th Annual Meeting of the Association for Computational Linguistics (Volume 1: Long Papers)}, pp.\  6327--6337, Dublin, Ireland, 2022{\natexlab{a}}. Association for Computational Linguistics.
\newblock \doi{10.18653/v1/2022.acl-long.438}.

\bibitem[Li et~al.(2022{\natexlab{b}})Li, Li, Xiong, and Hoi]{li2022blip}
Junnan Li, Dongxu Li, Caiming Xiong, and Steven C.~H. Hoi.
\newblock {BLIP:} bootstrapping language-image pre-training for unified vision-language understanding and generation.
\newblock In Kamalika Chaudhuri, Stefanie Jegelka, Le~Song, Csaba Szepesv{\'{a}}ri, Gang Niu, and Sivan Sabato (eds.), \emph{International Conference on Machine Learning, {ICML} 2022, 17-23 July 2022, Baltimore, Maryland, {USA}}, volume 162 of \emph{Proceedings of Machine Learning Research}, pp.\  12888--12900. {PMLR}, 2022{\natexlab{b}}.

\bibitem[Li et~al.(2019)Li, Yatskar, Yin, Hsieh, and Chang]{li2019visualbert}
Liunian~Harold Li, Mark Yatskar, Da~Yin, Cho-Jui Hsieh, and Kai-Wei Chang.
\newblock Visualbert: A simple and performant baseline for vision and language.
\newblock \emph{ArXiv preprint}, abs/1908.03557, 2019.

\bibitem[Li et~al.(2022{\natexlab{c}})Li, Lin, Zhang, Fu, Chen, Lou, and Chen]{li2022advance}
Yifei Li, Zeqi Lin, Shizhuo Zhang, Qiang Fu, Bei Chen, Jian-Guang Lou, and Weizhu Chen.
\newblock On the advance of making language models better reasoners.
\newblock \emph{ArXiv preprint}, abs/2206.02336, 2022{\natexlab{c}}.

\bibitem[Liu et~al.(2023)Liu, Li, Wu, and Lee]{liu2023visual}
Haotian Liu, Chunyuan Li, Qingyang Wu, and Yong~Jae Lee.
\newblock Visual instruction tuning.
\newblock \emph{ArXiv preprint}, abs/2304.08485, 2023.

\bibitem[Lu et~al.(2021)Lu, Qiu, Chen, Xia, Zhao, Zhang, Yu, Liang, and Zhu]{lu2021iconqa}
Pan Lu, Liang Qiu, Jiaqi Chen, Tony Xia, Yizhou Zhao, Wei Zhang, Zhou Yu, Xiaodan Liang, and Song-Chun Zhu.
\newblock Iconqa: A new benchmark for abstract diagram understanding and visual language reasoning.
\newblock In \emph{The 35th Conference on Neural Information Processing Systems (NeurIPS) Track on Datasets and Benchmarks}, 2021.

\bibitem[Lu et~al.(2022{\natexlab{a}})Lu, Mishra, Xia, Qiu, Chang, Zhu, Tafjord, Clark, and Kalyan]{lu2022learn}
Pan Lu, Swaroop Mishra, Tony Xia, Liang Qiu, Kai-Wei Chang, Song-Chun Zhu, Oyvind Tafjord, Peter Clark, and Ashwin Kalyan.
\newblock Learn to explain: Multimodal reasoning via thought chains for science question answering.
\newblock \emph{Advances in Neural Information Processing Systems}, 35:\penalty0 2507--2521, 2022{\natexlab{a}}.

\bibitem[Lu et~al.(2022{\natexlab{b}})Lu, Qiu, Chang, Wu, Zhu, Rajpurohit, Clark, and Kalyan]{lu2022dynamic}
Pan Lu, Liang Qiu, Kai-Wei Chang, Ying~Nian Wu, Song-Chun Zhu, Tanmay Rajpurohit, Peter Clark, and Ashwin Kalyan.
\newblock Dynamic prompt learning via policy gradient for semi-structured mathematical reasoning.
\newblock \emph{ArXiv preprint}, abs/2209.14610, 2022{\natexlab{b}}.

\bibitem[Lu et~al.(2022{\natexlab{c}})Lu, Qiu, Yu, Welleck, and Chang]{lu2022survey}
Pan Lu, Liang Qiu, Wenhao Yu, Sean Welleck, and Kai-Wei Chang.
\newblock A survey of deep learning for mathematical reasoning.
\newblock \emph{ArXiv preprint}, abs/2212.10535, 2022{\natexlab{c}}.

\bibitem[Lu et~al.(2023)Lu, Peng, Cheng, Galley, Chang, Wu, Zhu, and Gao]{lu2023chameleon}
Pan Lu, Baolin Peng, Hao Cheng, Michel Galley, Kai-Wei Chang, Ying~Nian Wu, Song-Chun Zhu, and Jianfeng Gao.
\newblock Chameleon: Plug-and-play compositional reasoning with large language models.
\newblock In \emph{The Thirty-seventh Conference on Neural Information Processing Systems (NeurIPS 2023)}, 2023.

\bibitem[Magister et~al.(2022)Magister, Mallinson, Adamek, Malmi, and Severyn]{magister2022teaching}
Lucie~Charlotte Magister, Jonathan Mallinson, Jakub Adamek, Eric Malmi, and Aliaksei Severyn.
\newblock Teaching small language models to reason.
\newblock \emph{ArXiv preprint}, abs/2212.08410, 2022.

\bibitem[OpenAI(2023)]{gpt4v}
OpenAI.
\newblock Gpt-4v(ision) system card, 2023.

\bibitem[Peng et~al.(2024)Peng, Wang, Dong, Hao, Huang, Ma, Ye, and Wei]{peng2024grounding}
Zhiliang Peng, Wenhui Wang, Li~Dong, Yaru Hao, Shaohan Huang, Shuming Ma, Qixiang Ye, and Furu Wei.
\newblock Grounding multimodal large language models to the world.
\newblock In \emph{The Twelfth International Conference on Learning Representations}, 2024.
\newblock URL \url{https://openreview.net/forum?id=lLmqxkfSIw}.

\bibitem[Radford et~al.(2021)Radford, Kim, Hallacy, Ramesh, Goh, Agarwal, Sastry, Askell, Mishkin, Clark, Krueger, and Sutskever]{radford2021learning}
Alec Radford, Jong~Wook Kim, Chris Hallacy, Aditya Ramesh, Gabriel Goh, Sandhini Agarwal, Girish Sastry, Amanda Askell, Pamela Mishkin, Jack Clark, Gretchen Krueger, and Ilya Sutskever.
\newblock Learning transferable visual models from natural language supervision.
\newblock In Marina Meila and Tong Zhang (eds.), \emph{Proceedings of the 38th International Conference on Machine Learning, {ICML} 2021, 18-24 July 2021, Virtual Event}, volume 139 of \emph{Proceedings of Machine Learning Research}, pp.\  8748--8763. {PMLR}, 2021.

\bibitem[Rae et~al.(2021)Rae, Borgeaud, Cai, Millican, Hoffmann, Song, Aslanides, Henderson, Ring, Young, Rutherford, Hennigan, Menick, Cassirer, Powell, Driessche, Hendricks, Rauh, Huang, Glaese, Welbl, Dathathri, Huang, Uesato, Mellor, Higgins, Creswell, McAleese, Wu, Elsen, Jayakumar, Buchatskaya, Budden, Sutherland, Simonyan, Paganini, Sifre, Martens, Li, Kuncoro, Nematzadeh, Gribovskaya, Donato, Lazaridou, Mensch, Lespiau, Tsimpoukelli, Grigorev, Fritz, Sottiaux, Pajarskas, Pohlen, Gong, Toyama, d'Autume, Li, Terzi, Mikulik, Babuschkin, Clark, Casas, Guy, Jones, Bradbury, Johnson, Hechtman, Weidinger, Gabriel, Isaac, Lockhart, Osindero, Rimell, Dyer, Vinyals, Ayoub, Stanway, Bennett, Hassabis, Kavukcuoglu, and Irving]{gopher}
Jack~W. Rae, Sebastian Borgeaud, Trevor Cai, Katie Millican, Jordan Hoffmann, Francis Song, John Aslanides, Sarah Henderson, Roman Ring, Susannah Young, Eliza Rutherford, Tom Hennigan, Jacob Menick, Albin Cassirer, Richard Powell, George van~den Driessche, Lisa~Anne Hendricks, Maribeth Rauh, Po-Sen Huang, Amelia Glaese, Johannes Welbl, Sumanth Dathathri, Saffron Huang, Jonathan Uesato, John Mellor, Irina Higgins, Antonia Creswell, Nat McAleese, Amy Wu, Erich Elsen, Siddhant Jayakumar, Elena Buchatskaya, David Budden, Esme Sutherland, Karen Simonyan, Michela Paganini, Laurent Sifre, Lena Martens, Xiang~Lorraine Li, Adhiguna Kuncoro, Aida Nematzadeh, Elena Gribovskaya, Domenic Donato, Angeliki Lazaridou, Arthur Mensch, Jean-Baptiste Lespiau, Maria Tsimpoukelli, Nikolai Grigorev, Doug Fritz, Thibault Sottiaux, Mantas Pajarskas, Toby Pohlen, Zhitao Gong, Daniel Toyama, Cyprien de~Masson d'Autume, Yujia Li, Tayfun Terzi, Vladimir Mikulik, Igor Babuschkin, Aidan Clark, Diego de~Las Casas, Aurelia Guy, Chris Jones,
  James Bradbury, Matthew Johnson, Blake Hechtman, Laura Weidinger, Iason Gabriel, William Isaac, Ed~Lockhart, Simon Osindero, Laura Rimell, Chris Dyer, Oriol Vinyals, Kareem Ayoub, Jeff Stanway, Lorrayne Bennett, Demis Hassabis, Koray Kavukcuoglu, and Geoffrey Irving.
\newblock Scaling language models: Methods, analysis \& insights from training gopher.
\newblock \emph{ArXiv preprint}, abs/2112.11446, 2021.

\bibitem[Raffel et~al.(2020)Raffel, Shazeer, Roberts, Lee, Narang, Matena, Zhou, Li, and Liu]{raffel2020exploring}
Colin Raffel, Noam Shazeer, Adam Roberts, Katherine Lee, Sharan Narang, Michael Matena, Yanqi Zhou, Wei Li, and Peter~J. Liu.
\newblock Exploring the limits of transfer learning with a unified text-to-text transformer.
\newblock \emph{J. Mach. Learn. Res.}, 21:\penalty0 140:1--140:67, 2020.

\bibitem[Reid et~al.(2024)Reid, Savinov, Teplyashin, Lepikhin, Lillicrap, Alayrac, Soricut, Lazaridou, Firat, Schrittwieser, et~al.]{reid2024gemini}
Machel Reid, Nikolay Savinov, Denis Teplyashin, Dmitry Lepikhin, Timothy Lillicrap, Jean-baptiste Alayrac, Radu Soricut, Angeliki Lazaridou, Orhan Firat, Julian Schrittwieser, et~al.
\newblock Gemini 1.5: Unlocking multimodal understanding across millions of tokens of context.
\newblock \emph{arXiv preprint arXiv:2403.05530}, 2024.

\bibitem[Rubin et~al.(2022)Rubin, Herzig, and Berant]{rubin2021learning}
Ohad Rubin, Jonathan Herzig, and Jonathan Berant.
\newblock Learning to retrieve prompts for in-context learning.
\newblock In \emph{Proceedings of the 2022 Conference of the North American Chapter of the Association for Computational Linguistics: Human Language Technologies}, pp.\  2655--2671, Seattle, United States, 2022. Association for Computational Linguistics.
\newblock \doi{10.18653/v1/2022.naacl-main.191}.

\bibitem[Schwenk et~al.(2022)Schwenk, Khandelwal, Clark, Marino, and Mottaghi]{schwenk2022okvqa}
Dustin Schwenk, Apoorv Khandelwal, Christopher Clark, Kenneth Marino, and Roozbeh Mottaghi.
\newblock A-okvqa: A benchmark for visual question answering using world knowledge.
\newblock In \emph{Computer Vision--ECCV 2022: 17th European Conference, Tel Aviv, Israel, October 23--27, 2022, Proceedings, Part VIII}, pp.\  146--162. Springer, 2022.

\bibitem[Taori et~al.(2023)Taori, Gulrajani, Zhang, Dubois, Li, Guestrin, Liang, and Hashimoto]{taori2023alpaca}
Rohan Taori, Ishaan Gulrajani, Tianyi Zhang, Yann Dubois, Xuechen Li, Carlos Guestrin, Percy Liang, and Tatsunori~B Hashimoto.
\newblock Alpaca: A strong, replicable instruction-following model.
\newblock \emph{Stanford Center for Research on Foundation Models. https://crfm. stanford. edu/2023/03/13/alpaca. html}, 2023.

\bibitem[Thoppilan et~al.(2022)Thoppilan, De~Freitas, Hall, Shazeer, Kulshreshtha, Cheng, Jin, Bos, Baker, Du, Li, Lee, Zheng, Ghafouri, Menegali, Huang, Krikun, Lepikhin, Qin, Chen, Xu, Chen, Roberts, Bosma, Zhao, Zhou, Chang, Krivokon, Rusch, Pickett, Srinivasan, Man, Meier-Hellstern, Morris, Doshi, Santos, Duke, Soraker, Zevenbergen, Prabhakaran, Diaz, Hutchinson, Olson, Molina, Hoffman-John, Lee, Aroyo, Rajakumar, Butryna, Lamm, Kuzmina, Fenton, Cohen, Bernstein, Kurzweil, Aguera-Arcas, Cui, Croak, Chi, and Le]{lamda}
Romal Thoppilan, Daniel De~Freitas, Jamie Hall, Noam Shazeer, Apoorv Kulshreshtha, Heng-Tze Cheng, Alicia Jin, Taylor Bos, Leslie Baker, Yu~Du, YaGuang Li, Hongrae Lee, Huaixiu~Steven Zheng, Amin Ghafouri, Marcelo Menegali, Yanping Huang, Maxim Krikun, Dmitry Lepikhin, James Qin, Dehao Chen, Yuanzhong Xu, Zhifeng Chen, Adam Roberts, Maarten Bosma, Vincent Zhao, Yanqi Zhou, Chung-Ching Chang, Igor Krivokon, Will Rusch, Marc Pickett, Pranesh Srinivasan, Laichee Man, Kathleen Meier-Hellstern, Meredith~Ringel Morris, Tulsee Doshi, Renelito~Delos Santos, Toju Duke, Johnny Soraker, Ben Zevenbergen, Vinodkumar Prabhakaran, Mark Diaz, Ben Hutchinson, Kristen Olson, Alejandra Molina, Erin Hoffman-John, Josh Lee, Lora Aroyo, Ravi Rajakumar, Alena Butryna, Matthew Lamm, Viktoriya Kuzmina, Joe Fenton, Aaron Cohen, Rachel Bernstein, Ray Kurzweil, Blaise Aguera-Arcas, Claire Cui, Marian Croak, Ed~Chi, and Quoc Le.
\newblock Lamda: Language models for dialog applications.
\newblock \emph{ArXiv preprint}, abs/2201.08239, 2022.

\bibitem[Vaswani et~al.(2017)Vaswani, Shazeer, Parmar, Uszkoreit, Jones, Gomez, Kaiser, and Polosukhin]{vaswani2017attention}
Ashish Vaswani, Noam Shazeer, Niki Parmar, Jakob Uszkoreit, Llion Jones, Aidan~N. Gomez, Lukasz Kaiser, and Illia Polosukhin.
\newblock Attention is all you need.
\newblock In Isabelle Guyon, Ulrike von Luxburg, Samy Bengio, Hanna~M. Wallach, Rob Fergus, S.~V.~N. Vishwanathan, and Roman Garnett (eds.), \emph{Advances in Neural Information Processing Systems 30: Annual Conference on Neural Information Processing Systems 2017, December 4-9, 2017, Long Beach, CA, {USA}}, pp.\  5998--6008, 2017.

\bibitem[Wang et~al.(2022{\natexlab{a}})Wang, Deng, and Sun]{wang2022iteratively}
Boshi Wang, Xiang Deng, and Huan Sun.
\newblock Iteratively prompt pre-trained language models for chain of thought.
\newblock In \emph{Proceedings of the 2022 Conference on Empirical Methods in Natural Language Processing}, pp.\  2714--2730, Abu Dhabi, United Arab Emirates, 2022{\natexlab{a}}. Association for Computational Linguistics.

\bibitem[Wang et~al.(2022{\natexlab{b}})Wang, Wei, Schuurmans, Le, Chi, and Zhou]{cot_wei_sc}
Xuezhi Wang, Jason Wei, Dale Schuurmans, Quoc Le, Ed~Chi, and Denny Zhou.
\newblock Self-consistency improves chain of thought reasoning in language models.
\newblock \emph{ArXiv preprint}, abs/2203.11171, 2022{\natexlab{b}}.

\bibitem[Wang et~al.(2022{\natexlab{c}})Wang, Wei, Schuurmans, Le, Chi, and Zhou]{wang2022rationale}
Xuezhi Wang, Jason Wei, Dale Schuurmans, Quoc Le, Ed~Chi, and Denny Zhou.
\newblock Rationale-augmented ensembles in language models.
\newblock \emph{ArXiv preprint}, abs/2207.00747, 2022{\natexlab{c}}.

\bibitem[Wei et~al.(2022{\natexlab{a}})Wei, Tay, Bommasani, Raffel, Zoph, Borgeaud, Yogatama, Bosma, Zhou, Metzler, Chi, Hashimoto, Vinyals, Liang, Dean, and Fedus]{wei2022emergent}
Jason Wei, Yi~Tay, Rishi Bommasani, Colin Raffel, Barret Zoph, Sebastian Borgeaud, Dani Yogatama, Maarten Bosma, Denny Zhou, Donald Metzler, Ed~H. Chi, Tatsunori Hashimoto, Oriol Vinyals, Percy Liang, Jeff Dean, and William Fedus.
\newblock Emergent abilities of large language models.
\newblock \emph{Transactions on Machine Learning Research}, 2022{\natexlab{a}}.
\newblock Survey Certification.

\bibitem[Wei et~al.(2022{\natexlab{b}})Wei, Wang, Schuurmans, Bosma, Chi, Le, and Zhou]{cot_wei}
Jason Wei, Xuezhi Wang, Dale Schuurmans, Maarten Bosma, Ed~Chi, Quoc Le, and Denny Zhou.
\newblock Chain of thought prompting elicits reasoning in large language models.
\newblock \emph{ArXiv preprint}, abs/2201.11903, 2022{\natexlab{b}}.

\bibitem[Wu et~al.(2021)Wu, Kong, Bi, Li, and Kao]{wu2021good}
Zhiyong Wu, Lingpeng Kong, Wei Bi, Xiang Li, and Ben Kao.
\newblock Good for misconceived reasons: An empirical revisiting on the need for visual context in multimodal machine translation.
\newblock In \emph{Proceedings of the 59th Annual Meeting of the Association for Computational Linguistics and the 11th International Joint Conference on Natural Language Processing (Volume 1: Long Papers)}, pp.\  6153--6166, Online, 2021. Association for Computational Linguistics.
\newblock \doi{10.18653/v1/2021.acl-long.480}.

\bibitem[Yasunaga et~al.(2022)Yasunaga, Aghajanyan, Shi, James, Leskovec, Liang, Lewis, Zettlemoyer, and Yih]{yasunaga2022retrieval}
Michihiro Yasunaga, Armen Aghajanyan, Weijia Shi, Rich James, Jure Leskovec, Percy Liang, Mike Lewis, Luke Zettlemoyer, and Wen-tau Yih.
\newblock Retrieval-augmented multimodal language modeling.
\newblock \emph{Proceedings of the 40th International Conference on Machine Learning, PMLR}, pp.\  39755--39769, 2022.

\bibitem[Yu et~al.(2019)Yu, Yu, Cui, Tao, and Tian]{yu2019mcan}
Zhou Yu, Jun Yu, Yuhao Cui, Dacheng Tao, and Qi~Tian.
\newblock Deep modular co-attention networks for visual question answering.
\newblock In \emph{{IEEE} Conference on Computer Vision and Pattern Recognition, {CVPR} 2019, Long Beach, CA, USA, June 16-20, 2019}, pp.\  6281--6290. Computer Vision Foundation / {IEEE}, 2019.
\newblock \doi{10.1109/CVPR.2019.00644}.

\bibitem[Yue et~al.(2024)Yue, Ni, Zhang, Zheng, Liu, Zhang, Stevens, Jiang, Ren, Sun, Wei, Yu, Yuan, Sun, Yin, Zheng, Yang, Liu, Huang, Sun, Su, and Chen]{yue2023mmmu}
Xiang Yue, Yuansheng Ni, Kai Zhang, Tianyu Zheng, Ruoqi Liu, Ge~Zhang, Samuel Stevens, Dongfu Jiang, Weiming Ren, Yuxuan Sun, Cong Wei, Botao Yu, Ruibin Yuan, Renliang Sun, Ming Yin, Boyuan Zheng, Zhenzhu Yang, Yibo Liu, Wenhao Huang, Huan Sun, Yu~Su, and Wenhu Chen.
\newblock Mmmu: A massive multi-discipline multimodal understanding and reasoning benchmark for expert agi.
\newblock In \emph{Proceedings of CVPR}, 2024.

\bibitem[Zhang et~al.(2023{\natexlab{a}})Zhang, Han, Zhou, Hu, Yan, Lu, Li, Gao, and Qiao]{zhang2023llama}
Renrui Zhang, Jiaming Han, Aojun Zhou, Xiangfei Hu, Shilin Yan, Pan Lu, Hongsheng Li, Peng Gao, and Yu~Qiao.
\newblock Llama-adapter: Efficient fine-tuning of language models with zero-init attention.
\newblock \emph{ArXiv preprint}, abs/2303.16199, 2023{\natexlab{a}}.

\bibitem[Zhang et~al.(2023{\natexlab{b}})Zhang, Li, Cui, Cai, Liu, Fu, Huang, Zhao, Zhang, Chen, et~al.]{zhang2023siren}
Yue Zhang, Yafu Li, Leyang Cui, Deng Cai, Lemao Liu, Tingchen Fu, Xinting Huang, Enbo Zhao, Yu~Zhang, Yulong Chen, et~al.
\newblock Siren's song in the ai ocean: A survey on hallucination in large language models.
\newblock \emph{arXiv preprint arXiv:2309.01219}, 2023{\natexlab{b}}.

\bibitem[Zhang et~al.(2020)Zhang, Chen, Wang, Utiyama, Sumita, Li, and Zhao]{zhang2020neural}
Zhuosheng Zhang, Kehai Chen, Rui Wang, Masao Utiyama, Eiichiro Sumita, Zuchao Li, and Hai Zhao.
\newblock Neural machine translation with universal visual representation.
\newblock In \emph{8th International Conference on Learning Representations, {ICLR} 2020, Addis Ababa, Ethiopia, April 26-30, 2020}. OpenReview.net, 2020.

\bibitem[Zhang et~al.(2023{\natexlab{c}})Zhang, Chen, Wang, Utiyama, Sumita, Li, and Zhao]{zhang2023universal}
Zhuosheng Zhang, Kehai Chen, Rui Wang, Masao Utiyama, Eiichiro Sumita, Zuchao Li, and Hai Zhao.
\newblock Universal multimodal representation for language understanding.
\newblock \emph{IEEE Transactions on Pattern Analysis and Machine Intelligence}, pp.\  1--18, 2023{\natexlab{c}}.
\newblock \doi{10.1109/TPAMI.2023.3234170}.

\bibitem[Zhang et~al.(2023{\natexlab{d}})Zhang, Zhang, Li, and Smola]{zhang2022automatic}
Zhuosheng Zhang, Aston Zhang, Mu~Li, and Alex Smola.
\newblock Automatic chain of thought prompting in large language models.
\newblock In \emph{The Eleventh International Conference on Learning Representations}, 2023{\natexlab{d}}.

\bibitem[Zhao et~al.(2023)Zhao, Cai, Si, Ma, An, Chen, Liu, Wang, Han, and Chang]{zhao2023mmicl}
Haozhe Zhao, Zefan Cai, Shuzheng Si, Xiaojian Ma, Kaikai An, Liang Chen, Zixuan Liu, Sheng Wang, Wenjuan Han, and Baobao Chang.
\newblock Mmicl: Empowering vision-language model with multi-modal in-context learning.
\newblock \emph{arXiv preprint arXiv:2309.07915}, 2023.

\bibitem[Zhou et~al.(2022)Zhou, Sch{\"a}rli, Hou, Wei, Scales, Wang, Schuurmans, Bousquet, Le, and Chi]{zhou2022least}
Denny Zhou, Nathanael Sch{\"a}rli, Le~Hou, Jason Wei, Nathan Scales, Xuezhi Wang, Dale Schuurmans, Olivier Bousquet, Quoc Le, and Ed~Chi.
\newblock Least-to-most prompting enables complex reasoning in large language models.
\newblock \emph{ArXiv preprint}, abs/2205.10625, 2022.

\bibitem[Zhu et~al.(2023)Zhu, Chen, Shen, Li, and Elhoseiny]{zhu2023minigpt}
Deyao Zhu, Jun Chen, Xiaoqian Shen, Xiang Li, and Mohamed Elhoseiny.
\newblock Minigpt-4: Enhancing vision-language understanding with advanced large language models.
\newblock In \emph{The Twelfth International Conference on Learning Representations}, 2023.

\end{thebibliography}
\bibliographystyle{tmlr}

\newpage
\appendix
\section{Extended Analysis for the Challenge of Multimodal-CoT}

\subsection{Additional Examples of Misleading through Hallucinated Rationales}\label{appendix:misleading}

\begin{figure*}[!htb]
  \begin{center}
   \includegraphics[width=1.0\textwidth]{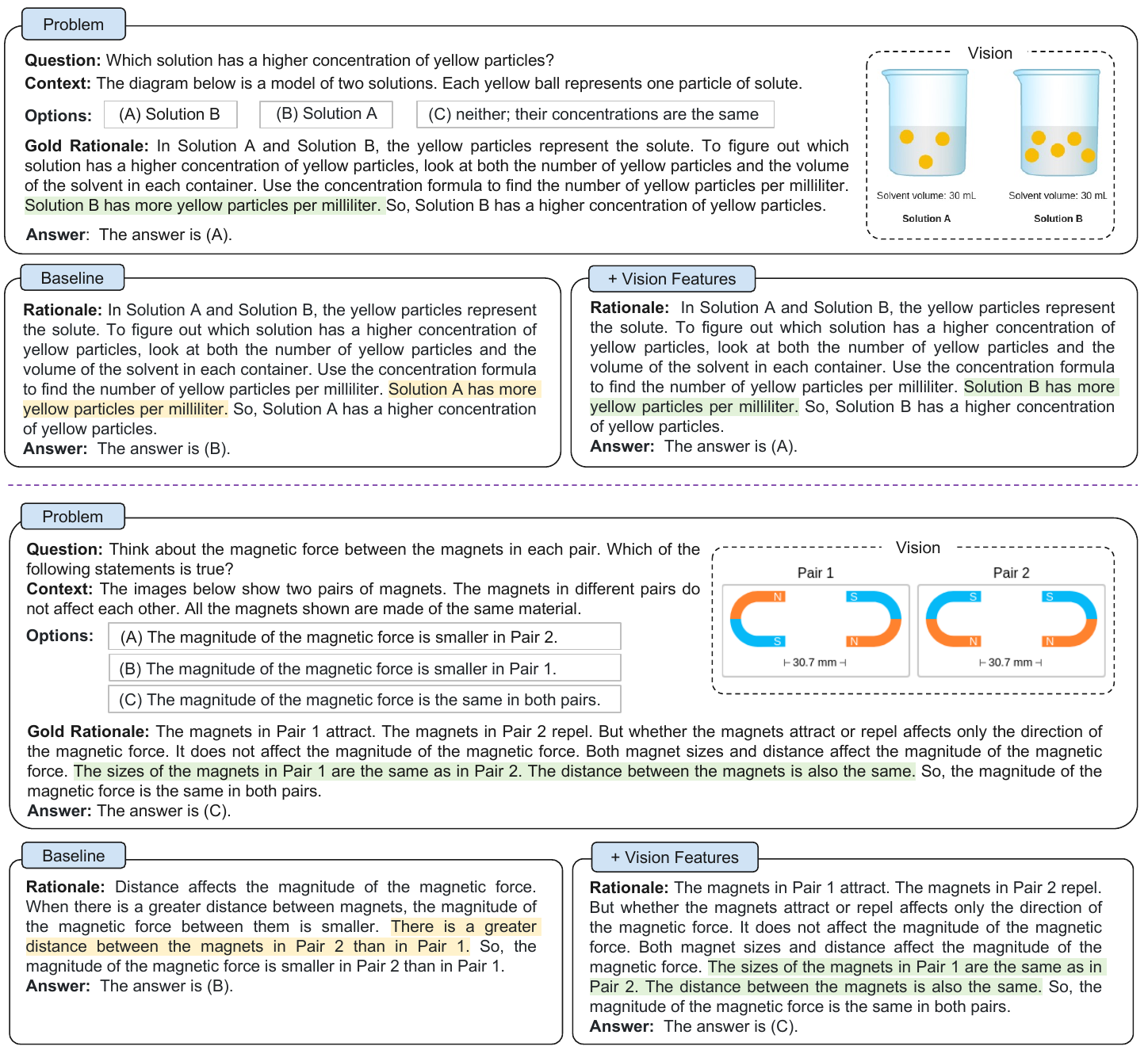}
  \end{center}
  \vspace{-3mm}
  \caption{Examples of the two-stage framework without vision features (baseline) and with vision features (ours) for generating rationales and predicting answers. The upper part presents the problem details, and the lower part shows the outputs of the baseline and our method. 
  }
    \vspace{-3mm}
  \label{fig_pre_case2}
\end{figure*}

Based on our case studies (Section 3.2), we have observed a tendency for the baseline model to generate hallucinated rationales. Here, we present additional examples to illustrate this phenomenon, as depicted in Figure \ref{fig_pre_case2}.

\subsection{Two-Stage Training Performance with Different Sizes of LMs}\label{appendix:lms}

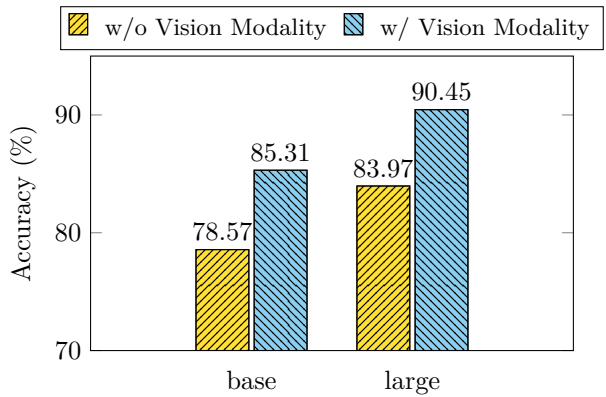
\begin{figure}[htb]
  \begin{center}
{
\pgfplotsset{compat=1.13,
    /pgfplots/ybar legend/.style={
    /pgfplots/legend image code/.code={%
       \draw[##1,/tikz/.cd,yshift=-0.25em]
        (0cm,0cm) rectangle (7pt,0.8em);},
   },
}
\pgfplotsset{width=8cm, height=5.5cm}
    \centering
    
    \begin{tikzpicture}  
        \begin{axis}  
        [  
            ybar,
            ymin=70, ymax=95,
            ytick={70,80,90,100},
            major x tick style = transparent,
            bar width=20pt,
            enlarge x limits=1.0,
            ylabel={Accuracy (\%)},
            symbolic x coords={0,1},  
            xtick=data,  
            xticklabels={base, large},
            nodes near coords,  
            nodes near coords align={vertical},  
        legend cell align=left,
         legend columns=2 row=1,
                legend style={
                        at={(0.5,1.02)},
                        anchor=south,
                        column sep=1ex,
                        font=\small,
                }
            ]  
        \addplot[ybar, fill=bananayellow,  postaction={pattern=north east lines}] coordinates {
            (0,78.57)(1, 83.97)
        };  
        \addplot[ybar, fill=babyblue,  postaction={pattern=north west lines}] coordinates {
            (0, 85.31)(1, 90.45)
        };
        \legend{w/o Vision Modality, w/ Vision Modality} 
        \end{axis}  
    \end{tikzpicture}
    \caption{
   Answer accuracy with different sizes of LMs.\label{fig_lm_size}
    }
}
  \end{center}
  \label{fig_scaling}
\end{figure}

In Section 3, we observed that the inclusion of vision features has a positive impact on the generation of more effective rationales, consequently resulting in improved answer accuracy. In addition to incorporating vision features, another approach to addressing the issue of incorrect rationales is to scale the size of the language model (LM). Figure \ref{fig_lm_size} showcases the answer accuracy achieved by our two-stage training framework, both with and without the integration of vision features. Notably, when employing a larger LM, the baseline accuracy (without vision features) experiences a significant enhancement. This finding suggests that scaling the LM size could potentially alleviate the problem of incorrect rationales. However, it is crucial to acknowledge that the performance still falls considerably short of utilizing vision features. This outcome further validates the effectiveness of our Multimodal-CoT methodology across varying LM sizes.

\subsection{Discussion of the Possible Paradigms to Achieve Multimodal-CoT}
\begin{figure}[htb]
  \begin{center}
   \includegraphics[width=0.88\textwidth]{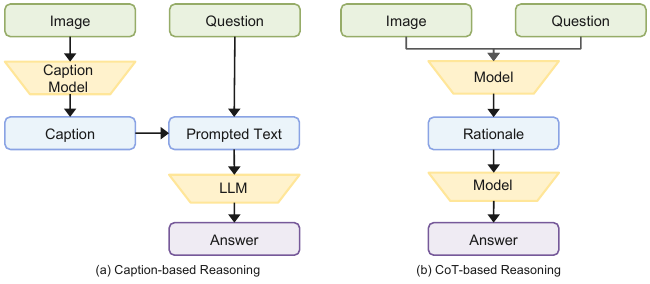}
  \end{center}
  \caption{Paradigms to achieve Multimodal-CoT.}
  \label{fig:paradigms}
\end{figure}

As discussed in Section 1, there are two primary approaches to facilitate Multimodal-CoT reasoning: (i) prompting LLMs and (ii) fine-tuning small models.
The common approach in the first approach is to unify the input from different modalities and prompt LLMs to perform reasoning  \citep{zhang2023llama,lu2023chameleon,liu2023visual,alayrac2022flamingo,hao2022language,yasunaga2022retrieval}. For instance, one way to achieve this is by extracting the caption of an image using a captioning model and then concatenating the caption with the original language input to feed LLMs. By doing so, visual information is conveyed to LLMs as text, effectively bridging the gap between modalities. This approach can be represented as the input-output format <image $\rightarrow$ caption, question + caption $\rightarrow$ answer>. We refer to this approach as \textbf{Caption-based Reasoning} (Figure \ref{fig:paradigms}a). It is worth noting that the effectiveness of this approach depends on the quality of the image caption, which may be susceptible to errors introduced during the transfer from image captioning to answer inference.

In contrast, an intriguing aspect of CoT is the ability to decompose complex problems into a series of simpler problems and solve them step by step. This transformation leads to a modification of the standard format <question $\rightarrow$ answer> into <question $\rightarrow$ rationale $\rightarrow$ answer>. Rationales, being more likely to reflect the reasoning processes leading to the answer, play a crucial role in this paradigm. Consequently, we refer to approaches following this paradigm as \textbf{CoT-based Reasoning}. The nomenclature has been widely adopted in the literature \citep{huang2022towards,zhang2022automatic,lu2022survey}.

Our work aligns with the paradigms of \textbf{CoT-based Reasoning} in the context of multimodal scenarios, specifically employing the <question + image $\rightarrow$ rationale $\rightarrow$ answer> framework (Figure \ref{fig:paradigms}b). This approach confers advantages on two fronts. Firstly, the Multimodal-CoT framework leverages feature-level interactions between vision and language inputs, enabling the model to gain a deeper understanding of the input information and facilitating more effective inference of answers by incorporating well-founded rationales. Our analysis has demonstrated that Multimodal-CoT offers notable benefits by mitigating hallucination and enhancing convergence, resulting in superior performance on our benchmark datasets. Secondly, the lightweight nature of Multimodal-CoT renders it compatible with resource constraints and circumvents any potential paywalls.

\section{Experimental Details}\label{app:exp}

\subsection{Details of Vision Features}\label{appendix:vision_features}
In Section 6.2, we compared four types of vision features, ViT \citep{dosovitskiy2021image}, CLIP \citep{radford2021learning}, DETR \citep{carion2020end}, and ResNet \citep{he2016deep}. The specific models are: (i) ViT: \textit{vit\_large\_patch32\_384},\footnote{\url{https://github.com/rwightman/pytorch-image-models}.} (ii) CLIP: RN101;\footnote{\url{https://github.com/jianjieluo/OpenAI-CLIP-Feature}.} (iii) DETR: \textit{detr\_resnet101\_dc5};\footnote{\url{https://github.com/facebookresearch/detr}.} (iv) ResNet: we use the averaged pooled features of a pre-trained ResNet50 CNN. 

Table \ref{tab:visual_dimension} presents the dimension of the vision features (after the function \textrm{VisionExtractor}($\cdot$) in Eq. 3). For ResNet-50, we repeat the 
pooled features of ResNet-50 to the same length as the text sequence to imitate the patch-like features, where each patch is the same as the pooled image features.

\begin{table}[htb]
    \centering
        \caption{Feature shape of vision features\label{tab:visual_dimension}}
\begin{tabular}{lc}\toprule
 {Method} & {Feature Shape} \\\midrule
 \quad  ViT & (145, 1024) \\
 \quad  CLIP & (49, 2048) \\
 \quad  DETR& (100, 256) \\
 \quad  ResNet & (512, 2048) \\
\bottomrule
\end{tabular}
\end{table}

\subsection{Datasets}
Our method is evaluated on the ScienceQA \citep{lu2022learn} and A-OKVQA \citep{schwenk2022okvqa} benchmark datasets. 

$\bullet$ ScienceQA is a large-scale multimodal science question dataset with annotated lectures and explanations. It contains 21$k$ multimodal multiple choice questions with rich domain diversity across 3 subjects, 26 topics, 127 categories, and 379 skills. The dataset is split into training, validation, and test splits with 12$k$, 4$k$, and 4$k$ questions, respectively. 

$\bullet$ A-OKVQA is a knowledge-based visual question answering benchmark, which has 25$k$ questions requiring a broad base of commonsense and world knowledge to answer. Each question is annotated with rationales that explain why a particular answer was correct according to necessary
facts or knowledge. It has 17$k$/1$k$/6$k$ questions for train/val/test. 

For ScienceQA, our model is evaluated on the test set. For A-OKVQA, our model is evaluated on the validation set as the test set is hidden.

\subsection{Implementation Details of Multimodal-CoT}
As the Multimodal-CoT task requires generating the reasoning chains and leveraging the vision features,  we adopt the T5 encoder-decoder architecture \citep{raffel2020exploring} under $\texttt{Base}$ (200M) and $\texttt{large}$ (700M) settings in our framework. We apply FLAN-Alpaca to initialize our model weights.\footnote{\url{https://github.com/declare-lab/flan-alpaca}.} We will show that Multimodal-CoT is generally effective with other backbone LMs, such as UnifiedQA \citep{khashabi2020unifiedqa} and FLAN-T5 \citep{chung2022scaling} (Section 6.1). The vision features are obtained by the frozen ViT-large encoder \citep{dosovitskiy2021image}. Since using image captions can slightly improve model performance, as shown in Section 3.3, we append the image captions to the context following \citet{lu2022learn}. The captions are generated by InstructBLIP \citep{instructblip}. We fine-tune the models up to 20 epochs, with a learning rate selected in \{5e-5, 8e-5\}. The maximum input sequence lengths for rationale generation and answer inference are 512 and 64, respectively. The batch size is 8. Our experiments are run on 8 NVIDIA Tesla V100 32G GPUs.

\section{Further Analysis}

\subsection{Examples of Rationale Generation with Large Models}\label{appendix:large}
\begin{figure}[htb]
  \begin{center}
    \vspace{-3mm}
   \includegraphics[width=0.98\textwidth]{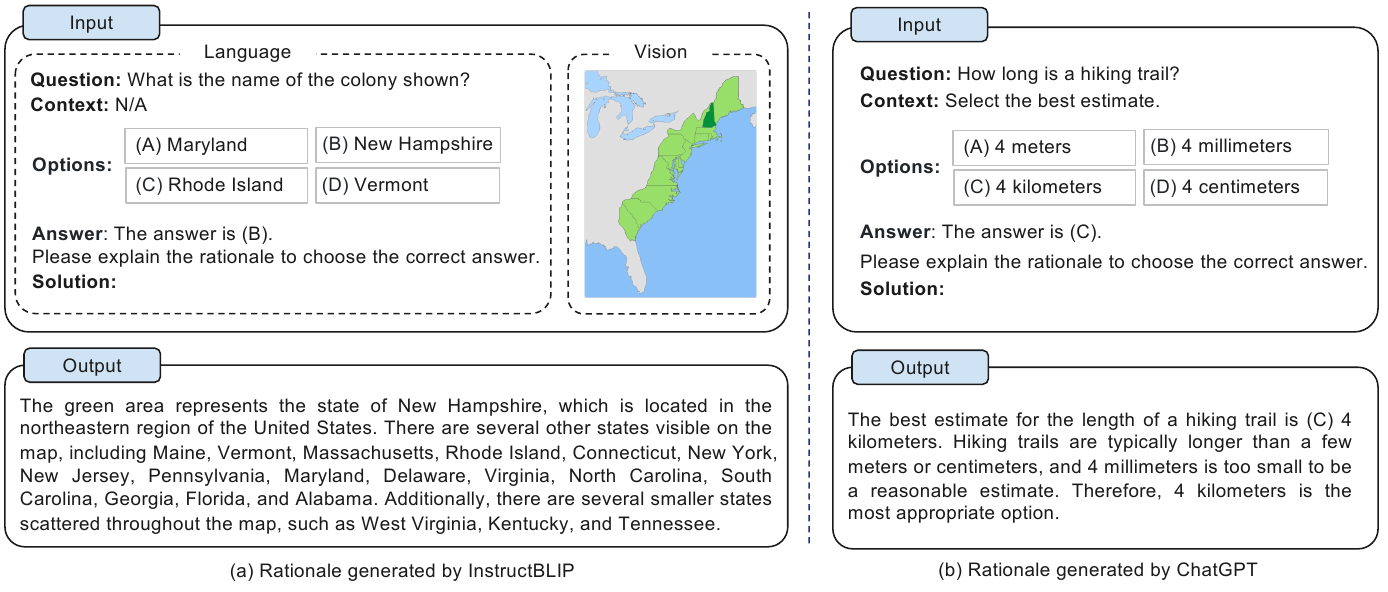}
  \end{center}
  \vspace{-3mm}
  \caption{Rationale generation examples.}
  \label{fig:rationale_generation}
\end{figure}

A recent flame is to leverage large language models or large vision-language models to generate reasoning chains for multimodal question answering problems \citep{zhang2023llama,lu2023chameleon,liu2023visual,alayrac2022flamingo,hao2022language,yasunaga2022retrieval}. We are interested in whether we can use large models to generate the rationales for Multimodal-CoT; thus breaking the need for datasets with human-annotated rationales. During the first-stage training of Multimodal-CoT, our target rationales are based on human annotation in the benchmark datasets. Now, we replace the target rationales with those generated by an LLM or a vision-language model. Concretely, we feed the questions with images (IMG) and the question without images (TXT) to InstructBLIP \citep{instructblip} (Figure \ref{fig:rationale_generation}a) and ChatGPT (Figure \ref{fig:rationale_generation}b) for zero-shot inference, respectively. Then, we use the generated pseudo-rationales as the target rationales for training instead of relying on the human annotation of reasoning chains.

\subsection{Detailed Results of Multimodal-CoT on Different Backbone Models}\label{app:backbone}

{To test the generality of the benefits of our approach to other backbone models, we alter the underlying LMs to other variants of different types. As detailed results shown in Table \ref{tab:apptable-backbones}, our approach is generally effective for the widely used backbone models.}

\begin{table*}[htb]
\centering
\caption{Detailed results of Multimodal-CoT on different backbone models. }
\small
\renewcommand\tabcolsep{5pt} 
\resizebox{1.0\linewidth}{!}
{
\begin{tabular}{l|cccccccc|l} 
\toprule
 Model & NAT & SOC & LAN & TXT & IMG & NO & G1-6 & G7-12 & ~Avg \\
\midrule
MM-CoT on UnifiedQA & 80.60 & 89.43 & 81.00 & 80.50 & 80.61 & 81.74 & 82.38 & 82.86 & 82.55 \\
MM-CoT on FLAN-T5 & 81.39 & 90.89 & 80.64 & 80.79 & 80.47 & 82.58 & 83.48 & 82.66 & 83.19 \\
 MM-CoT on FLAN-Alpaca & 84.06 & 92.35 & 82.18 & 82.75 & 82.75 & 84.74 & 85.79 & 84.44 & 85.31 \\
 \bottomrule
\end{tabular}
}
 \label{tab:apptable-backbones}
\end{table*}

\section{Examples of Case Studies}\label{appendix:case_study}
\begin{figure}[htb]
  \begin{center}
   \includegraphics[width=0.45\textwidth]{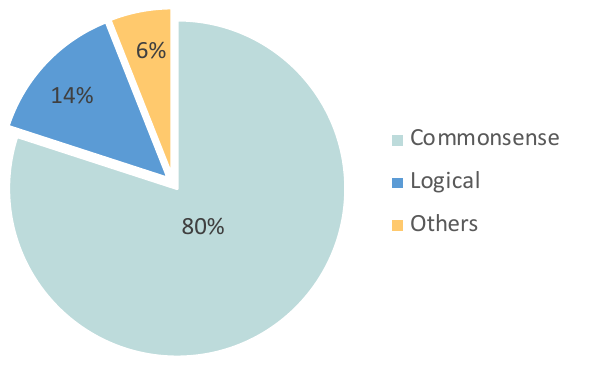}
  \end{center}
  \caption{Categorization analysis.}
  \label{fig-app:analysis_error}
\end{figure}

To gain deeper insights into the behavior of Multimodal-CoT and facilitate future research, we manually analyzed randomly selected examples generated by our approach. The categorization results are illustrated in Figure \ref{fig-app:analysis_error}. We examined 50 samples that yielded incorrect answers and categorized them accordingly. 

\begin{figure*}[htb]
  \begin{center}
   \includegraphics[width=1\textwidth]{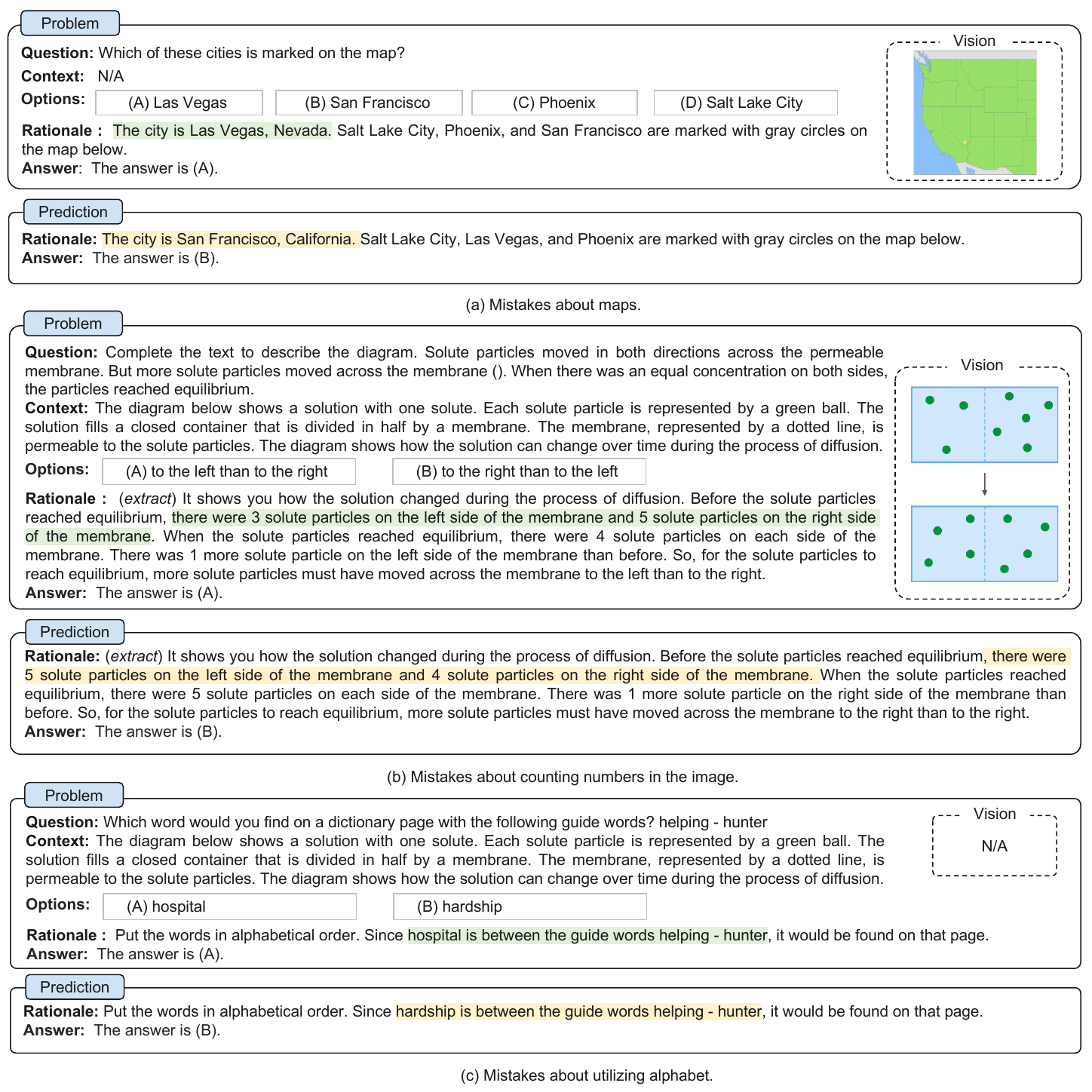}
  \end{center}
  \caption{Examples of commonsense mistakes.}
  \label{fig-case-study-incorrect-fact}
\end{figure*}

The most prevalent error type is commonsense mistakes, accounting for 80\% of the errors. These mistakes occur when the model is faced with questions that require commonsense knowledge, such as interpreting maps (Figure \ref{fig-case-study-incorrect-fact}a), counting objects in images (Figure \ref{fig-case-study-incorrect-fact}b), or utilizing the alphabet (Figure \ref{fig-case-study-incorrect-fact}c). 

\begin{figure*}[!htb]
  \begin{center}
   \includegraphics[width=1\textwidth]{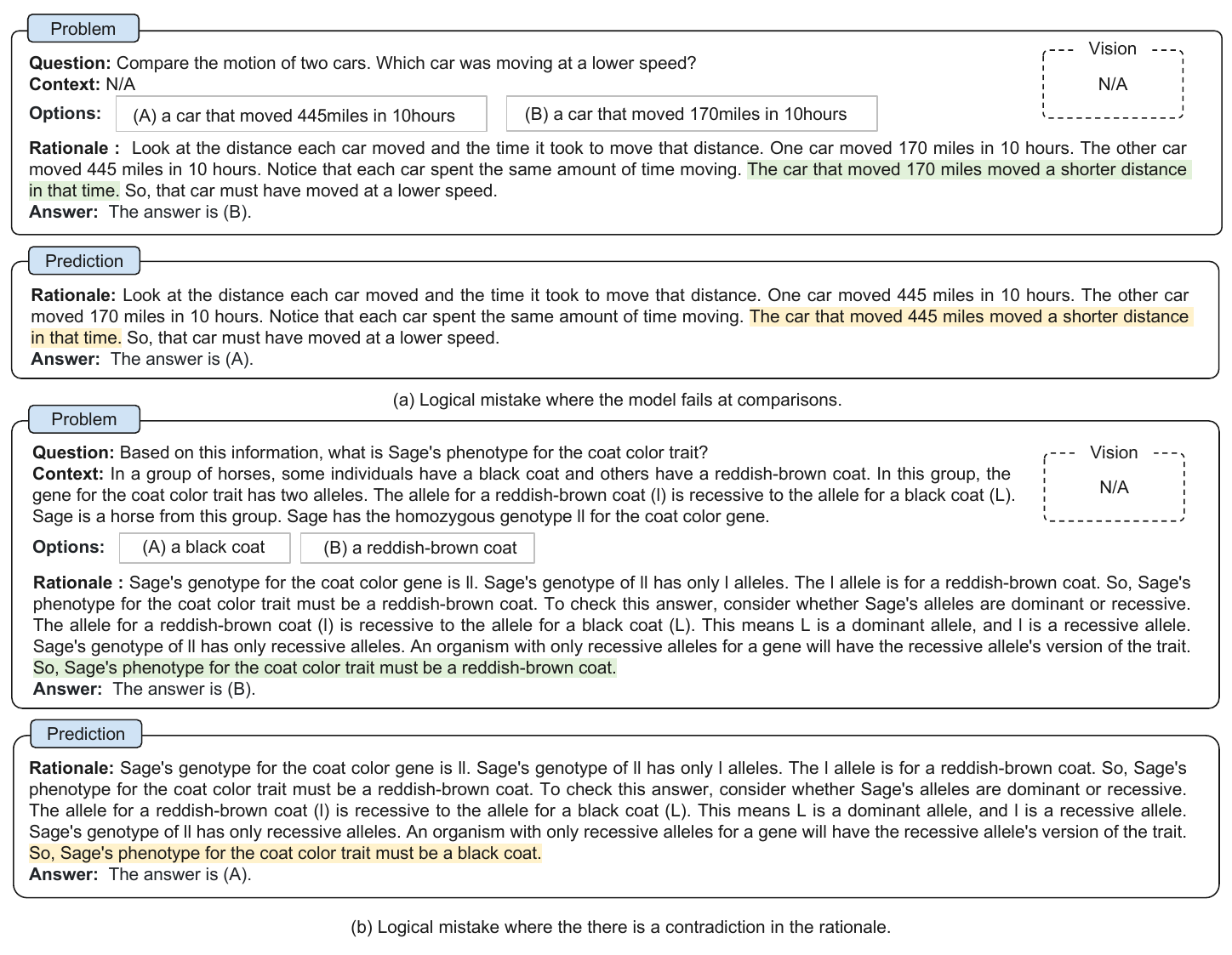}
  \end{center}
  \caption{Examples of logical mistakes.}
  \label{fig-case-study-incorrect-logical}
\end{figure*}

The second error type is logical mistakes, constituting 14\% of the errors, which involve comparison mistakes (Figure \ref{fig-case-study-incorrect-logical}a) and contradictions in the reasoning process (Figure \ref{fig-case-study-incorrect-logical}b).

\begin{figure*}[!htb]
  \begin{center}
   \includegraphics[width=1\textwidth]{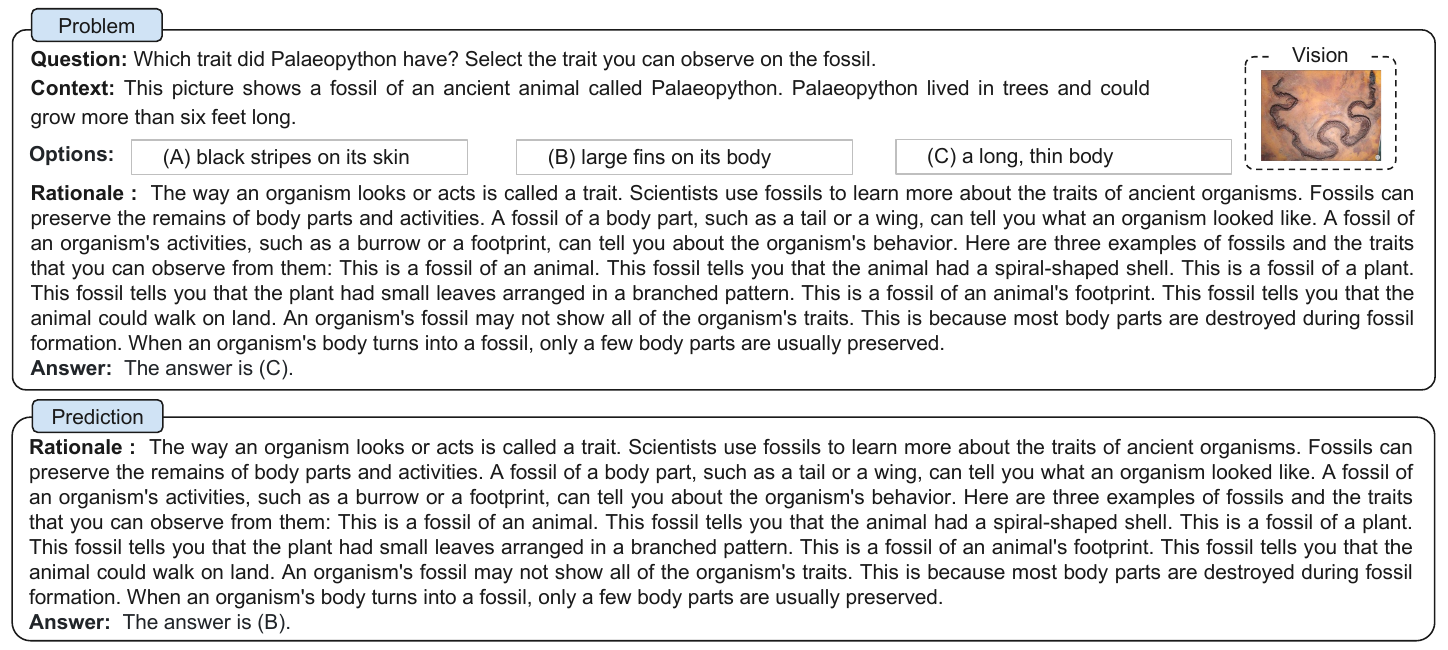}
  \end{center}
  \caption{Examples of answers are incorrect while the CoT is correct.}
  \label{fig-case-study-incorrect-correct}
\end{figure*}

Additionally, we have observed cases where incorrect answers are provided despite the CoT being either empty or correct, amounting to 6\% of the errors. The CoT in these cases may not necessarily influence the final answer (Figure \ref{fig-case-study-incorrect-correct}).


The analysis reveals potential avenues for future research. Enhancements can be made to Multimodal-CoT by: (i) integrating more informative visual features and strengthening the interaction between language and vision to enable comprehension of maps and numerical counting; (ii) incorporating commonsense knowledge; and (iii) implementing a filtering mechanism, such as using only relevant CoTs to infer answers and disregarding irrelevant ones.

\end{document}